\documentclass[lettersize,journal]{IEEEtran}
\usepackage{times}
\usepackage{soul}
\usepackage{url}
\usepackage[hidelinks]{hyperref}
\usepackage[utf8]{inputenc}
\usepackage[small]{caption}
\usepackage{amsmath}
\usepackage{amsthm}
\usepackage{booktabs}
\usepackage{algorithm}
\usepackage[switch]{lineno}
\usepackage{float}
\usepackage{wrapfig}
\usepackage{graphicx}
\usepackage{subcaption}
\usepackage{amsmath,amsfonts}
\usepackage{algorithmic}
\usepackage{algorithm}
\usepackage{array}
\usepackage{textcomp}
\usepackage{stfloats}
\usepackage{url}
\usepackage{verbatim}
\usepackage{graphicx}
\usepackage{cite}
\usepackage{xcolor}
\usepackage{cleveref}
\newcommand{\mathbold}[1]{\ensuremath{\boldsymbol{\mathbf{#1}}}}

\newcommand{\g}{\,|\,}

\newcommand{\nestedmathbold}[1]{{\mathbold{#1}}}

\newcommand{\mbw}{\nestedmathbold{w}}
\newcommand{\mbx}{\nestedmathbold{x}}
\newcommand{\mby}{\nestedmathbold{y}}

\newcommand{\mbH}{\nestedmathbold{H}}

\newcommand{\mbX}{\nestedmathbold{X}}

\newcommand{\mbphi}{\nestedmathbold{\phi}}

\newcommand{\mbtheta}{\nestedmathbold{\theta}}

\DeclareMathOperator*{\argmax}{arg\,max}

\newcommand{\cD}{\mathcal{D}}

\newcommand{\cN}{\mathcal{N}}

\begin{document}

\title{FedSI: Federated Subnetwork Inference \\for Efficient Uncertainty Quantification}

\author{Hui Chen, Hengyu Liu, Zhangkai Wu, Xuhui Fan, and Longbing Cao,~\IEEEmembership{Senior Member,~IEEE}
\thanks{The work is partially sponsored by Australian Research Council Discovery, LIEF and Future Fellowship grants (DP190101079, DP240102050, FT190100734 and FT190100734).}
\thanks{Hui Chen, Xuhui Fan and Longbing Cao are with School of Computing, Macquarie University, Australia. (email: hui.chen2@students.mq.edu.au, xuhui.fan@mq.edu.au, longbing.cao@mq.edu.au).
Hengyu Liu is with Aalborg University, Aalborg 9220, Denmark. (email: heli@cs.aau.dk).
Zhangkai Wu is with University of Technology Sydney, Australia. (email: berenwu1938@gmail.com).
}}

\maketitle

\begin{abstract}
While deep neural networks (DNNs) based personalized federated learning (PFL) is demanding for addressing data heterogeneity and shows promising performance, existing methods for federated learning~(FL) suffer from efficient systematic uncertainty quantification. The Bayesian DNNs-based PFL is usually questioned of either over-simplified model structures or high computational and memory costs. In this paper, we introduce FedSI, a novel Bayesian DNNs-based subnetwork inference PFL framework. FedSI is  simple and scalable by leveraging Bayesian methods to incorporate systematic uncertainties effectively. It implements a client-specific subnetwork inference mechanism, selects network parameters with large variance to be inferred through posterior distributions, and fixes the rest as deterministic ones. FedSI achieves fast and scalable inference while preserving the systematic uncertainties to the fullest extent. Extensive experiments on three different benchmark datasets demonstrate that FedSI outperforms existing Bayesian and non-Bayesian FL baselines in heterogeneous FL scenarios.
\end{abstract}


\section{Introduction}
\IEEEPARstart{F}{ederated} leaning (FL) \cite{cao2023bayesian,li2022evolutionary,guo2023gpt4graph,zhao2022personalized,tan2022towards} has emerged as a critical methodology to meet the increasing modeling demand on decentralized applications \cite{Cao_DeAI} with data privacy concerns \cite{mcmahan2017communication}. An FL system is usually composed of a central server and several local clients, with the server training a single global model and with each client retaining its data locally to maintain privacy throughout the learning process~\cite{mcmahan2017communication}. Previous studies show that the above framework can obtain good performance when the local data across clients are i.i.d.~(i.e., independent and identically distributed) \cite{t2020personalized,achituve2021personalized,yao2022trajgat,du2022understanding}. However, in real application scenarios,  data distributions among clients may be significantly different due to the distinctions between their specific preferences, causing non-IID data with uncertainties such as data drift and concept drift \cite{C22beyiid}. Consequently, a single global model over all the clients could lead to deviated generalization errors~\cite{chen2021bridging,xu2023personalized,li2020compression,li2021trace}. In order to address the data heterogeneity, various Personalized Federated Learning (PFL) methods, in particular those based on Deep Neural Networks (DNNs) \cite{arivazhagan2019federated,collins2021exploiting,oh2021fedbabu,cheng2017survey}, multi-task learning \cite{smith2017federated}, and meta learning \cite{fallah2020personalized}, have been proposed to construct a customized model for each client.

While these PFL approaches demonstrate remarkable performance in the heterogeneous data environment, most of them are unable to quantify uncertainties, which is particularly crucial for safety-critical applications \cite{amodei2016concrete,kokolakis2022safety,liu2023dynamic,jiang2024reinforcement}. Moreover, performance degradation on limited data is another challenge for these approaches, especially for large-scale DNNs models~\cite{kairouz2021advances,du2022gbk,liu2024probabilistic,yu2023subspace}. By contrast, Bayesian DNNs-based PFL algorithms can quantify uncertainties and obtain performance gain on limited data \cite{cao2023bayesian}. Bayesian DNNs-based PFL approaches usually use stochastic (e.g., Stochastic Gradient Langevin Dynamics (SGLD) \cite{welling2011bayesian,fan2021continuous,fan2021poisson}) or deterministic (e.g., Mean-Field Variational Inference (MFVI) \cite{rodriguez2022function,fan2023free,fan2020bayesian,wu2024weakly}) schemes to approximate the posterior distributions over model parameters, requiring a large amount of sampling steps or invoking over-simplified independence assumptions over random variables. That is, such Bayesian DNNs settings often lead to \textbf{substantial computational and memory cost} or \textbf{degraded model performance}. 

Recently, there has been substantial studies on Bayesian DNNs ~\cite{izmailov2020subspace,sharma2023bayesian,daxberger2021bayesian}, showing that these Bayesian DNNs issues can be mitigated by replacing  the posterior distribution approximation over the whole network parameters with that on a subnetwork. Inspired by this approach, we propose a framework of personalized \textbf{Fed}erated learning with \textbf{S}ubnetwork \textbf{I}nference (\textbf{FedSI}) to perform posterior inference over an individual subset of network parameters for each client, while keeping other parameters deterministic. In particular, we first decouple the entire Bayesian neural network into two parts: a \emph{representation layer} to learn common features and a \emph{decision layer} to make personalized classification decision, drawing from conventional DNNs-based PFL methods~\cite{oh2021fedbabu,collins2021exploiting,pillutla2022federated}. Then, we perform subnetwork inference on each client and local updating on the representation layers, while leaving the decision layer updating out of the entire federated training process. We prove that FedSI actually makes posterior inference for network parameters with high variance and makes point estimations for those with low variance. Consequently, FedSI reduces the number of posterior distributions for network parameters while preserving uncertainties to the maximum extent possible.


Specifically, in the local update stage for each client, we first fix the deterministic parameters of the classifier layer, which is randomly initialized at the beginning of federated training, and then calculate the Maximum A Posterior (MAP) estimate over the parameters from representation layers. We then utilize the Linearized Laplace Approximation (LLA) to infer a full-covariance Gaussian posterior over the subnetwork of representation layers. In the aggregation stage, through the incorporation of degenerate Gaussian distributions and continuous learning fashion, we  dynamically transform and combine the deterministic representation parameters and stochastic representation parameters. Consequently, the server  receives all updated  distribution parameters of representation layers from participating clients and averages them for learning common subnetwork representation information. 

\textbf{Main Contributions.} Our main contributions are summarized as follows:

\begin{itemize}

\item We propose a novel Bayesian PFL framework called FedSI, which can achieve scalable inference and preserve network uncertainty at the same time. This introduced client-specific subnetwork inference method may help mitigate distribution drift resulting from data heterogeneity during training, thereby enhancing the convergence speed of  modeling.

\item  To the best of our knowledge,  FedSI is the first work to investigate efficient uncertainty quantification in the context of FL. Thanks to the low-dimensional subnetworks within the representation layer, our model is able to significantly reduce computational consumption while achieving efficient inference. More importantly, we overcome the independence assumption over model parameters and provide a full-covariance structure for them which aligns more closely with real-world scenarios.


\item We design a flexible global aggregation strategy which effectively transforms and combines deterministic representation parameters and stochastic representation parameters, and accelerates model convergence with continual learning paradigms.

\item We conduct extensive experiments on three public benchmark datasets with diversified experimental settings. The experiment results demonstrate the superiority of FedSI compared with other baselines, including Bayesian and non-Bayesian FL methods.
\end{itemize}

The remainder of this paper is organized as follows. We give a brief overview of FL and LLA in Section \ref{section 2}. Section \ref{section 3} presents the proposed FedSI. In Section \ref{section 5}, we give the experimental design and corresponding experimental results. The related work is provided in Section \ref{section 6}, Finally, Section \ref{section 7} concludes this paper.

\section{Preliminaries}
In this section, we describe the relevant background about FL and LLA, which will lay the foundation for our proposed FedSI.

\subsection{Federated Learning}
\label{section 2}
We consider a typical FL system with $N$ clients and a central server, where each client $i$ has a training dataset $\mathcal{D}_i=\{(\mathbf{x}_{i,j},y_{i,j})\}_{j=1}^{n_i}$, with $(\mbx_{i,j}, y_{i,j})$ being the $j$-th feature-label pair and with $n_i$ being the number of data points in client $i$, the underlying optimization goal can be formulated as:
\begin{equation}
\label{1}
\begin{aligned}
\min_{\{\mbw_i\}_i}\left\{F(\{\mbw_i\}_i):=\frac{1}{N} \sum_{i=1}^N \left(F_i(\mbw_i)+\frac{\lambda}{2} \|\mbw_i-\mbw\|^2\right)\right\}.
\end{aligned}
\end{equation}
where $\{\mbw_i\}_i$ is the set of local model parameters for all clients, and $\mbw$ represents the global model parameter for the server, respectively. $F_i(\mbw_i):=\mathbb{E}_{(\mathbf{x},y)\sim P(\cD_i)}[\mathcal{L}(\mbw_i;\mathbf{x},y)]$ denotes the expected loss of one data point from client $i$, where $\mathcal{L}(\mbw_i;\mathbf{x},y)$ stands for the loss function on one feature-label pair $(\mbx, y)$, given the model parameter $\mbw_i$. $\lambda$ is the regularization coefficient to control the degree of personalization.

A generic FL setting aims to find a single global model across all the clients (i.e., $\mbw=\mbw_1=\cdots=\mbw_N$), which can cope with homogeneous data environment across clients. Nonetheless, local clients often suffer from distribution drift, leading to challenges for a single model to generalize well across all clients. To this end, PFL methods deal with this issue through designing personalized local model $\mbw_i$ for each client, which can fit client-specific data distribution while learning some common knowledge under a federated manner \cite{kotelevskii2022fedpop}. Furthermore, Eq.~\eqref{1} shows the nature of personalization in existing PFL is to seek a balance between the global model and personalized local models over local updating.

\subsection{Linearized Laplace Approximation~(LLA)}
\label{section 2-B}

LLA is a prominent approximate inference method for its exceptional performance compared to alternative methods \cite{foong2019between,immer2021improving,daxberger2021bayesian}. Bayesian Neural Networks~(BNNs) in a supervised learning task are used to specify the details of LLA. Given feature matrix $\mathbf{X} \in \mathbb{R}^{n \times I} $, label vector $\mathbf{y} \in \mathbb{R}^O$, and model parameter $\mbw \in \mathbb{R}^P$, using $f_{\mbw}(\mathbf{x})$ to represent BNN's output for $\mbx$, we can write the likelihood term as $p(\mby\g f_{\mbw}(\mbX))$.  Given dataset $\cD$, the target is to learn the posterior distribution over the model parameters $\mbw$:
\begin{equation} \label{eq:formal-posterior-of-w}
p(\mbw | D)=p(\mathbf{y} | \mathbf{X}) \propto p(\mathbf{y} | f_{\mbw}(\mathbf{X})) p(\mbw),
\end{equation}
where $p(\mbw)$ is the prior distribution over $\mbw$ and is set as an isotropic multivariate Gaussian distribution $p(\mbw)=\mathcal{N}(\mbw|\mathbf{0}, \alpha \mathbf{I})$ in this work. Based on Eq.~\eqref{eq:formal-posterior-of-w}, $\mbw$ can be estimated through MAP estimation to maximize its posterior distribution $p(\mbw|\cD)$ as: 
\begin{equation}
\mbw_{\mathrm{MAP}}=\argmax _{\mbw}[\log p(\mathbf{y} | f_{\mbw}(\mathbf{X}))+\log p(\mbw)].
\end{equation}

Given $\mbw_{\mathrm{MAP}}$, the LLA uses a second-order Taylor expansion $q(\mbw)$ around $\mbw_{\mathrm{MAP}}$ \cite{bishop2006pattern} to approximate the posterior distribution $p(\mbw|\mathcal{D})$ as:
\begin{align} 
\label{eq:w-taylor-series}
&\log p(\mbw |\mathcal{D}) \nonumber \\
&\simeq \log p(\mbw_{\mathrm{MAP}} | \mathcal{D})-\frac{(\mbw-\mbw_{\mathrm{MAP}})^{\mathrm{T}} \mathbf{H}(\mbw-\mbw_{\mathrm{MAP}})}{2}\\
&=\log q(\mbw), \nonumber
\end{align}
where 
$\mathbf{H}=- \nabla^2_{\mbw\mbw} \log p(\mathbf{y} | f_{\mbw}(\mathbf{X}))|_{\mbw=\mbw_{\mathrm{MAP}}}+\alpha^{-1} \mathbf{I} 
$.


Since $\mbw$ only appears with a quadratic format in Eq.~(\ref{eq:w-taylor-series}), $q(\mbw)$ can be formulated as a Gaussian distribution with full-covariance:
\begin{align} \label{eq:q_w}
{q(\mbw)}&=\frac{|\mathbf{H}|^{\frac{1}{2}}}{(2 \pi)^{\frac{P}{2}}} \exp (-\frac{\left(\mbw-\mbw_{\mathrm{MAP}}\right)^{\mathrm{T}} \mathbf{H}\left(\mbw-\mbw_{\mathrm{MAP}}\right)}{2})\nonumber \\
&=\mathcal{N}\left(\mbw | \mbw_{\mathrm{MAP}}, \mathbf{H}^{-1}\right).
\end{align}

Usually  the \textit{generalized Gauss-Newton} (GGN) is used to approximate the computationally intractable Hessian Matrix $\mbH$ in Eq.~(\ref{eq:q_w}) as \cite{martens2020new}:
\begin{equation}
\label{eq:GGN}
\mathbf{H} \simeq \widehat{\mathbf{H}}= \mathbf{J}^{\mathrm{T}} \mathbf{\Lambda} \mathbf{J}+\alpha^{-1} \mathbf{I},
\end{equation}
where $\mathbf{J}=\nabla_{\mbw}f_{\mbw}(\mbx)\in \mathbb{R}^{O \times P}$ denotes the Jacobian matrix of model output over the parameters, and $\mathbf{\Lambda}=-\nabla^2_{ff} \log p(\mathbf{y}|f_{\mbw}(\mbx)) \in \mathbb{R}^{O \times O}$ stands for the noise precision. More details can be found in  Appendix.

Recently, Immer et al. \cite{immer2021improving} revealed that transforming the underlying probability model of a BNN into a Bayesian Generalized Linear Model (GLM) by approximating the BNN with a first-order Taylor expansion around the mode $\mbw_\mathrm{MAP}$ can effectively mitigate model underfitting. Therefore, we can write the corresponding linearized model as follows:
\begin{equation}
f_{\mathrm{lin}}(\mathbf{x},\mbw)= f(\mathbf{x},\mbw_{\mathrm{MAP}})+\widetilde{\mathbf{J}}(\mbw-\mbw_{\mathrm{MAP}}),
\end{equation}
where $\widetilde{\mathbf{J}}=\nabla_{\mbw_{\mathrm{MAP}}}f(\mathbf{x},\mbw_{\mathrm{MAP}})$ is a $O \times P$ Jacobian matrix.

To assume that, given the value $\mathbf{x}$, the corresponding $y$ follows a Gaussian distribution with mean $f_{\mbw}(\mbx)$ and variance $\epsilon$, i.e., $p(y|\mathbf{x},\mbw,\epsilon)=\mathcal{N}(y|f_{\mathrm{lin}}(\mathbf{x},\mbw),\epsilon)$. To this end, the predictive distribution of regression task can be performed analytically by a Gaussian of the form:
\begin{equation}
\begin{aligned}
p(y|\mathbf{x}, \mathcal{D})&= \int p(y|f_{\mathrm{lin}}(\mathbf{x},\mbw))p(\mbw|\mathcal{D})\mathrm{d}\mbw\\
&=\mathcal{N}(y|f(\mathbf{x},\mbw_{\mathrm{MAP}}),\mathbf{\Sigma}(\mathbf{x})+\epsilon),
\end{aligned}
\end{equation}
where $\mathbf{\Sigma}(\mathbf{x})=\widetilde{\mathbf{J}}^{\mathrm{T}}\widehat{\mathbf{H}}^{-1}\widetilde{\mathbf{J}}$. 

For classification task, we can obtain similar result with a categorical distribution $p(\mathbf{y}|\mathbf{x},\mbw)=\mathrm{Cat}(\mathbf{y}|\sigma(f_{\mathrm{lin}}(\mathbf{x},\mbw)))$. However, the predictive distribution cannot be evaluated analytically due to the logistic function $\sigma(\cdot)$. Here, we make use of the probit function $\Phi(\cdot)$, which has similar properties to the logistic function, to obtain a good analytic approximate solution \cite{barber1998ensemble,bishop2006pattern, fan2018rectangular}. Specifically, we show that
\begin{equation}
p(\mathbf{y}|\mathbf{x}, \mathcal{D}) \simeq \mathrm{Cat} \left(\mathbf{y}|\Phi\left(\kappa(\mathbf{\Sigma}(\mathbf{x}))f(\mathbf{x},\mbw_{\mathrm{MAP}})\right)\right),
\end{equation}
where $\kappa(\mathbf{\Sigma}(\mathbf{x}))=(1+\pi \mathrm{diag}(\mathbf{\Sigma}(\mathbf{x}))/8)^{-1/2}$.

\section{FedSI: Personalized Federated Learning with Subnetwork Inference}
\label{section 3}
In this section, we provide a detailed description of our framework FedSI, which proposes LLA with subnetwork selection of Bayesian neural networks to achieve accurate and efficient uncertainty quantification.

\subsection{Subnetwork Inference in Bayesian Neural Networks}

In line with the standard configuration of DNNs-based federated learning frameworks, as detailed in the prior literature \cite{arivazhagan2019federated,collins2021exploiting}, we further decouple the layers of BNNs into two parts: representation layers $R$ and decision layers $B$, parameterized by ${\boldsymbol{\theta}} \in \mathbb{R}^{|R|}$ and ${\boldsymbol{\phi}} \in \mathbb{R}^{|B|}$,  respectively. Consequently, the model parameter for client $i$ is denoted as $\mbw_i = ({\mbtheta}_i, {\mbphi}_i)$. Within our framework, we update the distributions of model parameters over the representation layers $R$ and send the updated distributions to the server for global aggregation throughout the entire training procedure, whereas the model parameters of the decision layers $B$ are updated by fine-tuning during the evaluation phase.


Nevertheless, it is hard to obtain the posterior distribution $p({\mbtheta}|\mathcal{D}_i)$ due to the nonlinearity of neural networks and high dimensionality of  model parameter ${\mbtheta}_i$. Therefore, existing DNNs-based PFL methods must resort to  various  techniques to approximate the true posterior distribution. These approximation techniques can be divided into two categories: deterministic and stochastic techniques. The former, such as MFVI,  learns an approximate posterior distribution with factorized assumptions over model parameters ${\mbtheta}_i$. The latter, such as SGLD, attempts to sample from the target posterior distribution $p({\mbtheta}_i|\mathcal{D}_i)$. However, it has been shown that the above approaches incur either significant computational and memory cost or poor model performance \cite{ovadia2019can,fort2019deep,foong2020expressiveness}. To tackle these issues, we introduce the LLA with subnetwork inference to BNNs. This process is formalized as follows.

\textbf{Step 1: MAP over Model Parameter ${\mbtheta}_i$.} We compute $\mbtheta_i$'s MAP estimation via gradient decent algorithms \cite{lecun2015deep}:
\begin{equation}
{\mbtheta}_{i,\mathrm{MAP}}=\argmax_{{\mbtheta}_i}[\log p(\mathbf{y} | \mathbf{X}, {\mbtheta}_i,{\mbphi}_i)+\log p({\mbtheta})].
\end{equation}
Here, we follow the continual learning manner by replacing the local prior distribution $p({\mbtheta}_i)$ with global distribution $p({\mbtheta})$ from the server, which can avoid catastrophic forgetting occurring between consecutive communication rounds and expedite the convergence process.

\textbf{Step 2: Subnetwork Selection.} The LLA algorithm, which has demonstrated remarkable effectiveness in making prediction \cite{foong2019between,immer2021improving}, is adopted here.  To begin, we employ the GGN method as outlined in Section \ref{section 2-B} to approximate the ground-truth posterior distribution over ${\mbtheta}_i$ as follows:
\begin{equation} \label{eq:client-approx-q}
 p({\mbtheta}_i|\mathcal{D}_i) 
\simeq {q_{R}(\mbtheta_i)}=\mathcal{N}\left({\mbtheta}_i | {\mbtheta}_{i,\mathrm{MAP}}, \widehat{\mathbf{H}}_i^{-1}\right),
\end{equation}
where $\widehat{\mathbf{H}}_i=\mathbf{J}_i^{\mathrm{T}} \mathbf{\Lambda}_i \mathbf{J}_i+\mathrm{diag}(\mathbf{g}^{-1})\in \mathbb{R}^{|R|\times|R|}$,  $\mathbf{J}_i=\nabla_{{\mbtheta}_i}f(\mathbf{x},{\mbtheta}_i,{\mbphi}_i)\in \mathbb{R}^{O \times|R|}$, $\mathbf{\Lambda}_i=-\nabla^2_{ff} \log p(\mathbf{y}|f(\mathbf{x},{\mbtheta}_i,{\mbphi}_i)) \in \mathbb{R}^{O \times O}$, and $\mathbf{g}\in \mathbb{R}^{|R|}$ is a vector whose $r$-th element $\sigma^2_{r}$ is the variance of $\theta_{r}$'s global distribution.

Unfortunately, computing and storing the complete $|R| \times |R|$ covariance matrix $\widehat{\mbH}$ in Eq.~\eqref{eq:client-approx-q} is computationally infeasible.

To this end, we propose to identify a subnetwork $S \subset R$ ($|S| \ll |R|$) within the representation layer $R$. The parameters  ${\boldsymbol{\theta}}_{i,S} \in \mathbb{R}^{|S| \times 1}$ within this subnetwork are still learned through its approximated posterior distribution. Meanwhile, the parameters outside  $S$ are determined via their MAP estimation. That is, we may express the posterior distribution of $\boldsymbol{\theta}_i$ as:
 \begin{align}
\label{14}
q_{S}({\mbtheta}_{i})&=q\left({\mbtheta}_{i,S}\right) \prod_{d\in R\backslash S} \delta\left({\mbtheta}_{i,d}-{\mbtheta}_{i,d,\mathrm{MAP}}\right)\nonumber \\
& = \cN(\mbtheta_{i,S}\g\mbtheta_{i,S,\text{MAP}}, \mbH_{i,S}^{-1})\prod_{d\in R\backslash S} \delta\left({\mbtheta}_{i,d}-{\mbtheta}_{i,d,\mathrm{MAP}}\right),
\end{align}
where $q\left({\mbtheta}_{i,S}\right)$ is the approximate posterior distribution of subnetwork $S$, and $\delta(\cdot)$ represents the Dirac delta function; ${\mbtheta}_{i,d}$ denotes a deterministic parameter of the representation layers. Here, $\widehat{\mathbf{H}}_{i,S}=\mathbf{J}_{i,S}^{\mathrm{T}} \mathbf{\Lambda}_i \mathbf{J}_{i,S}+\mathrm{diag}(\mathbf{g}^{-1}_{S})$, where $\mathbf{J}_{i,S}=\nabla_{{\mbtheta}_{i,S}}f(\mathbf{x},{\mbtheta}_i,{\mbphi}_i)\in \mathbb{R}^{O \times S}$, and $\mathbf{g}_{S}$ is a vector whose $s$-th element is $\sigma^2_{s}$, which is the global distribution's variance of $\theta_{s}$, $s=1,\cdots,S$.

The discrepancy between GGN-approximated posterior distribution $q_{R}({\mbtheta}_i)$ and its subnetwork-induced approximation $q_{S}({\mbtheta}_i)$ can be measured by the squared 2-Wasserstein distance~\cite{daxberger2021bayesian,chizat2020faster}, which is:
\begin{equation}
\label{16}
\begin{aligned}
&W_2(q_{R}({\mbtheta}_i),q_S({\mbtheta}_{i}))^2 \\
&=\operatorname{Tr}\left(\widehat{\mathbf{H}}_i^{-1}+\widehat{\mathbf{H}}_{i,S^*}^{-1}-2\sqrt{\widehat{\mathbf{H}}_{i,S}^{-1/2} \widehat{\mathbf{H}}_i^{-1} \widehat{\mathbf{H}}_{i,S^*}^{-1/2}}\right),
\end{aligned}
\end{equation}
where $\widehat{\mathbf{H}}_{i,S^*}^{-1}$ is equivalent to padding $\widehat{\mathbf{H}}_{i,S}^{-1}$ with zeros at the positions of the remaining deterministic parameters ${\mbtheta}_{i,D} \in \mathbb{R}^{R-S}$, ensuring that the shapes of $\widehat{\mathbf{H}}_{i,S^*}^{-1}$ and $\widehat{\mathbf{H}}_i^{-1}$ are identical. 


Further assuming model parameters $\mbtheta_{i, \mathcal{S}}$ are independent, Eq.~\eqref{16} can be simplified as:
\begin{equation}
\label{17}
W_2(q_{R}({\mbtheta}_i),q_{S}({\mbtheta}_{i}))^2 \approx \sum_{r=1}^{R} \sigma^2_{r} (1-\chi_{{\mbtheta}_{i,S}} ({\mbtheta}_{i,r})),
\end{equation}
where $\sigma^2_{r}$ denotes the marginal variance of the $r$-th parameter in the representation layer. 

The subnetwork $S$ is identified by minimizing the approximated squared 2-Wasserstein distance, denoted as Eq.~\eqref{17}. This minimization process effectively selects parameters characterized by the \emph{highest variances} for inclusion in $S$, as determined by the values of ${\sigma_r^2}_{r=1}^{|R|}$. It's important to note that our assumption of independence is specific to the selection of the subnetwork $S$ and doesn't affect the posterior inference, ensuring that the predictive performance remains unaffected. Further details on the main results can be found in  Appendix.

\textbf{Step 3: Posterior Inference.} After determining the subnetwork $\mathcal{S}$, the full-covariance Gaussian posterior over parameters ${\mbtheta}_{i,S}$ can be inferred via GGN-Laplace approximation:
\begin{equation}
\label{18}
\begin{aligned}
q({\mbtheta}_{i,S})=\mathcal{N}\left({\mbtheta}_{i,S} | {\mbtheta}_{i,S,\mathrm{MAP}}, \widehat{\mathbf{H}}_{i,S}^{-1}\right).
\end{aligned}
\end{equation}

Then, we can obtain the corresponding approximate posterior distribution according to Eq.~\eqref{14}. Therefore, it means that, on the representation layers, we only perform the Laplace inference over the subnetwork's stochastic parameters ${\mbtheta}_{i,S}$ while the remaining deterministic parameters ${\mbtheta}_{i,D}$ are fixed at the MAP values. Additionally, since the dimensions of the subnetwork's parameters is small, it becomes viable to store and invert $\widehat{\mathbf{H}}_{i,S}$. Specifically, by substituting full-covariance matrix $\widehat{\mathbf{H}}_{i}$ for sub-covariance matrix $\widehat{\mathbf{H}}_{i,S}$, the complexity of storing and inverting decreases from $O(R^2)$ and $O(R^3)$ to $O(S^2)$ and $O(S^3)$ respectively. 

\begin{figure*}[t]
    \centering
    \includegraphics[width=0.85\textwidth]{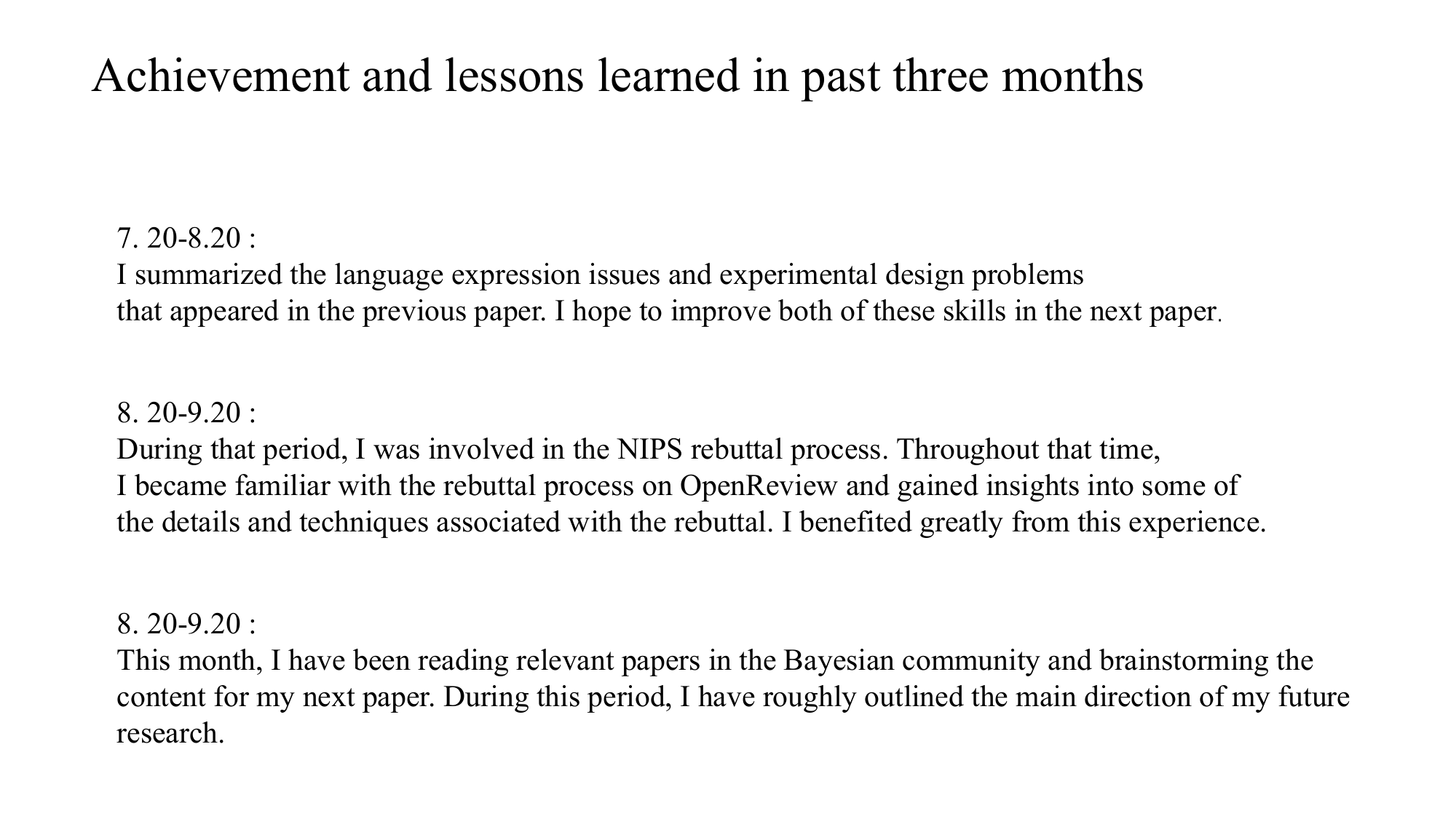}
    \caption{Personalized Federated Learning with Subnetwork Inference. After obtaining the MAP values, each client identifies its own subnetwork ${\mbtheta}_{i,S}$ and obtains its corresponding full-covariance Gaussian posterior $q({\mbtheta}_{i,S})$ through subnetwork inference (SI). Note that the decision parameters ${\mbphi}_i$ are fixed at their initial random values during this phase. Then, clients send the distribution parameters of representation parameters $\boldsymbol{\mu}^{t+1}_{\theta_i}$, $\boldsymbol{\sigma}^{t+1}_{\theta_i}$ to the server, which averages them to compute the distribution parameters of common representation parameters $\boldsymbol{\mu}^{t+1}_{\theta}$, $\boldsymbol{\sigma}^{t+1}_{\theta}$ for the next communication round. }
    \label{fig1}
\end{figure*}

For each communication round throughout the entire federated training process, we first follow  Steps 1-3 as previously described to  update the distributions of representation parameters ${\mbtheta}_i$, while the decision parameters ${\mbphi}_i$ are fixed at their deterministic initial random values during this phase. Subsequently, we send the updated distributions of representation parameters ${\mbtheta}_i$ to the server for aggregation. We show the process of such distribution training in Fig. \ref{fig1}.

\textbf{Step 4: Prediction on Test Dataset.} 
The uncertainty in prediction is calculated by defining the predictive distribution for new data sample as
\begin{equation}
p(y|\mathbf{x},\mathcal{D}_i)=\int p(y|\mathbf{x},{\mbtheta}_i)p({\mbtheta}|\mathcal{D}_i)\mathrm{d}{\mbtheta}_i.
\end{equation}

Once the entire federated training process is finished, each client can obtain the updated distributions of common representation parameters ${\mbtheta}$ from the server. We can  then only perform two steps to make prediction on the test dataset.

(1) Choose $S$ parameters with the highest variances as subnetwork's parameters and infer the full-covariance Gaussian posterior over parameters ${\mbtheta}_{i,S}$ by replacing the ${\mbtheta}_{i,\mathrm{MAP}}$ with  vector $\mathbf{h}$, whose $r$-th element is $\mu_{r}$, which is the mean of the global distribution $\theta_{r}$.

(2) Obtain the personalized parameters ${\mbphi}_{i,fin}$ by fine-tuning on the local dataset. 

We can rewrite the linearized model as the following:
\begin{equation}
\begin{aligned}
&f_{\mathrm{lin}}(\mathbf{x},\mathbf{h},{\mbphi}_{i,fin})\\
&=f(\mathbf{x},\mathbf{h},{\mbphi}_{i,fin})+\widetilde{\mathbf{J}}_{i,S}({\mbtheta}_{i,S}-\mathbf{h}_S),
\end{aligned}
\end{equation}
where $\widetilde{\mathbf{J}}_{i,S}=\nabla_{\mathbf{h}_S}f(\mathbf{x},\mathbf{h},{\mbphi}_{i,fin})$ is a $O \times S$ Jacobian matrix; $\mathbf{h}_S$ stand for the
mean vector of subnetwork.

Further, the predictive distribution for regression  and classification tasks can be expressed as
\begin{equation}
p(y|\mathbf{x}, \mathcal{D}_i)=\mathcal{N}(y|f(\mathbf{x},\mathbf{h},{\mbphi}_{i,fin}),\mathbf{\Sigma}_{i,S}(\mathbf{x})+\epsilon))
\end{equation}
and
\begin{equation}
p(\mathbf{y}|\mathbf{x}, \mathcal{D}_i) \simeq \mathrm{Cat} \left(\mathbf{y}|\Phi\left(\kappa(\mathbf{\Sigma}_{i,s}(\mathbf{x}))f(\mathbf{x},\mathbf{h},{\mbphi}_{i,fin})\right)\right),
\end{equation}
respectively. Here, $\mathbf{\Sigma}_{i,S}(\mathbf{x})=\widetilde{\mathbf{J}}_{i,S}^{\mathrm{T}}\widehat{\mathbf{H}}^{-1}\widetilde{\mathbf{J}}_{i,S}$ and $\kappa(\mathbf{\Sigma}_{i,S}(\mathbf{x}))=(1+\pi \mathrm{diag}(\mathbf{\Sigma}_{i,S}(\mathbf{x}))/8)^{-1/2}$.

\subsection{The FedSI Algorithm}
FedSI learns a common representation layer through alternating steps of local updating and global aggregation. In this part, we describe the entire training procedure of FedSI in detail.

\textbf{Local Updating.} For the local model training of the $i$-th client at the communication round $t$, we begin by substituting the prior distribution $p({\mbtheta}^t_i)$ with the global distribution $p({\mbtheta}^t)$ to obtain the MAP estimate ${\mbtheta}^t_{i,\mathrm{MAP}}$ over the representation parameters ${\mbtheta}_i$. Then, we select $S$ parameters with the highest variances  as the subnetwork's parameters by minimizing the squared 2-Wasserstein distance between $p({\mbtheta}^t_i|\mathcal{D}_i)$ and $q_S({\mbtheta}^t_{i})$. Finally, we compute the full-covariance Gaussian posterior over the subnetwork $q({\mbtheta}^t_{i,s})$ using the GGN-Laplace approximation. Note that during the entire local updating stage, personalized parameters ${\mbphi}^0_i$ remain consistently fixed at their deterministic initial random values.

\textbf{Global Aggregation.} Following a cross-device FL setting, a random subset $C^t$ of all clients participate in each communication round. When the local updating with respect to representation parameters ${\mbtheta}_i^t$ has been finished, the server then receives the updated distribution parameters of stochastic parameters $\boldsymbol{\mu}^{t+1}_{{\mbtheta}_{i,S}}$, $\boldsymbol{\sigma}^{t+1}_{{\mbtheta}_{i,S}}$  and deterministic parameters $\boldsymbol{\mu}^{t+1}_{{\mbtheta}_{i,D}}$, $\boldsymbol{\sigma}^{t+1}_{{\mbtheta}_{i,D}}$ from each participating client and performs the model averaging on all participating clients. It is noteworthy that, when conducting this averaging process, the element of deterministic parameters ${\mbtheta}^{t+1}_{i,D}$ can be considered as following a degenerate Gaussian distribution (with a variance of zero).

\begin{figure}[t]
    \centering
    \includegraphics[width=0.5\textwidth]{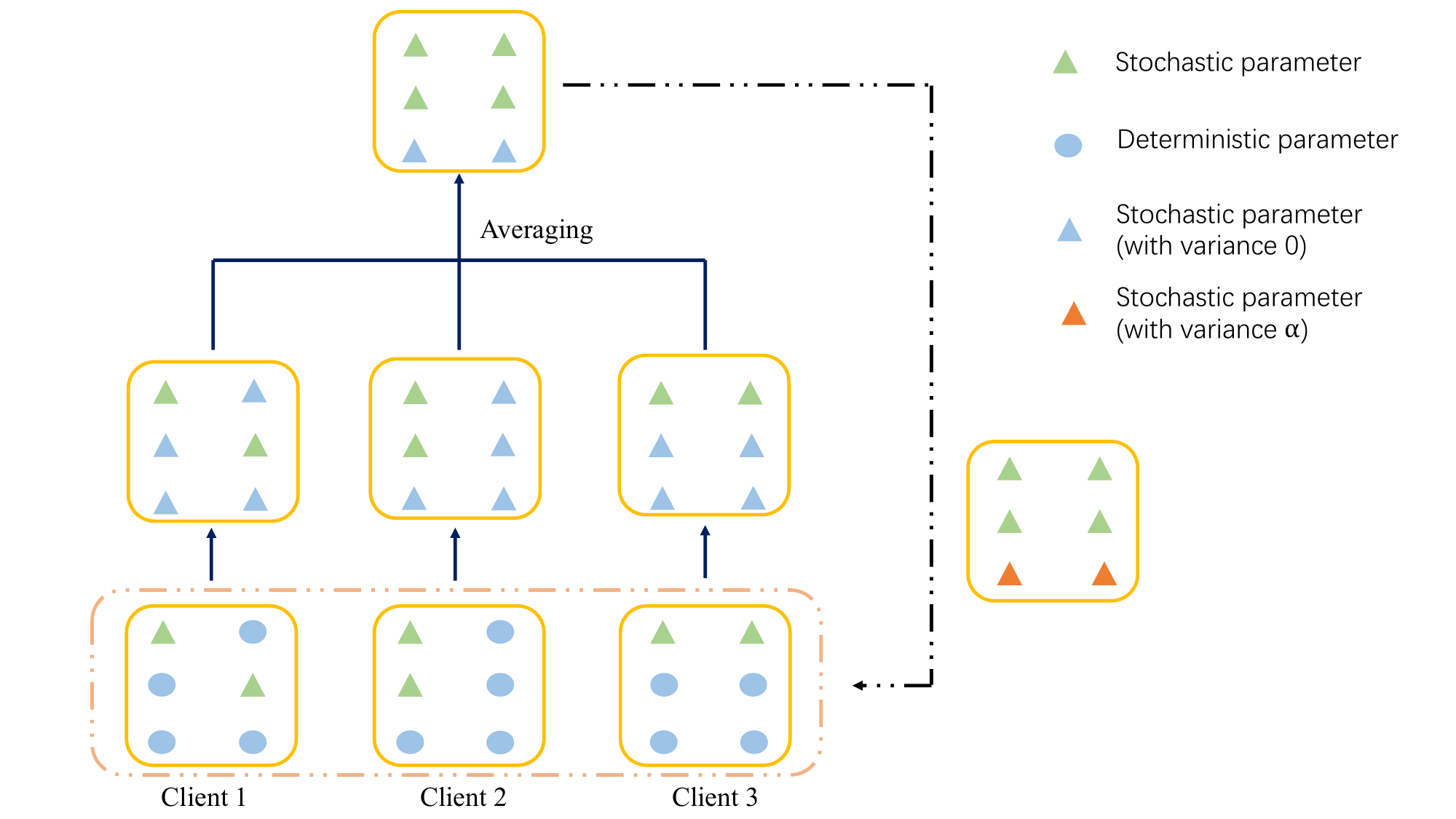}
    \caption{Global aggregation. \textbf{Before averaging:} Each participating client sends the server with the updated distribution parameters of stochastic parameters $\boldsymbol{\mu}^{t+1}_{{\mbtheta}_{i,S}}$, $\boldsymbol{\sigma}^{t+1}_{{\mbtheta}_{i,S}}$  and deterministic parameters $\boldsymbol{\mu}^{t+1}_{{\mbtheta}_{i,D}}$, $\boldsymbol{\sigma}^{t+1}_{{\mbtheta}_{i,D}}$. The element of ${\mbtheta}^{t+1}_{i,D}$ can be regarded as a degenerate Gaussian distribution for model averaging. \textbf{After averaging:} The server obtains the distribution parameters of stochastic common representation parameters $\boldsymbol{\mu}^{t+1}_{{\mbtheta}_S}$, $\boldsymbol{\sigma}^{t+1}_{{\mbtheta}_S}$ and deterministic common representation parameters $\boldsymbol{\mu}^{t+1}_{{\mbtheta}_D}$, $\boldsymbol{\sigma}^{t+1}_{{\mbtheta}_D}$. Note that the ${\mbtheta}^{t+1}_D$ is transformed into stochastic parameters following a Gaussian distribution with a covariance matrix of $\alpha \mathbf{I}$.}
    \label{fig2}
\end{figure}

After averaging, we obtain the distribution parameters of stochastic common representation parameters $\boldsymbol{\mu}^{t+1}_{{\mbtheta}_S}$, $\boldsymbol{\sigma}^{t+1}_{{\mbtheta}_S}$ and deterministic common representation parameters $\boldsymbol{\mu}^{t+1}_{{\mbtheta}_D}$, $\boldsymbol{\sigma}^{t+1}_{{\mbtheta}_D}$, which suggests that our global model retains the subnetwork selection preferences of each participating local client and learns the common subnetwork representation information. Note that, to construct the prior distribution for Step 1 in Section \ref{section 3}, we follow the general setting to convert the deterministic common representation parameters ${\mbtheta}^{t+1}_D$ into the stochastic ones, which follow a Gaussian distribution with a covariance matrix of $\alpha \mathbf{I}$. Fig. \ref{fig2} illustrates the above global aggregation process. The corresponding algorithm is formally presented in Algorithm \ref{alg1}.

\begin{algorithm}
	\renewcommand{\algorithmicrequire}{\textbf{Server executes:}}
	\renewcommand{\algorithmicensure}{\textbf{Client} Update($i, \boldsymbol{\mu}^t_{{\mbtheta}}, \boldsymbol{\sigma}^t_{{\mbtheta}}, {\mbphi}_i^0$)$\boldsymbol{:}$}
	\caption{\texttt{FedSI}: Personalized Federated Learning with Subnetwork Inference Algorithm}
	\label{alg1}
	\begin{algorithmic}
	\STATE \textbf{Input:} $T$-communication rounds, $K$-local epochs, $C$-random subset of all clients. Initialize: $\boldsymbol{\mu}^0_{{\mbtheta}}$, $\boldsymbol{\sigma}^0_{{\mbtheta}}$, ${\mbphi}^0_i$
       \REQUIRE
	\FOR{$t=0,1,\dots, T-1$}
        \STATE Sample $\mathcal{C}^t$ clients with size $C$ uniformly at random
        \FOR{each client $i \in \mathcal{C}^t$ \textbf{in parallel}}
        \STATE $\boldsymbol{\mu}^{t+1}_{{\mbtheta}_i}, \boldsymbol{\sigma}^{t+1}_{{\mbtheta}_i} \gets$ Client Update($i, \boldsymbol{\mu}^t_{{\mbtheta}}, \boldsymbol{\sigma}^t_{{\mbtheta}}, {\mbphi}_i^0$)
        \ENDFOR
        \STATE$\boldsymbol{\mu}^{t+1}_{{\mbtheta}}=\frac{1}{{C}}\sum_{i \in \mathcal{C}^t} \boldsymbol{\mu}^{t+1}_{{\mbtheta}_i}$, $\boldsymbol{\sigma}^{t+1}_{{\mbtheta}}=\frac{1}{{C}}\sum_{i \in \mathcal{C}^t} \boldsymbol{\sigma}^{t+1}_{{\mbtheta}_i}$
        \ENDFOR
        \ENSURE
        \STATE $\boldsymbol{\mu}^t_{{\mbtheta}_{i,0}} \gets \boldsymbol{\mu}^t_{{\mbtheta}}$, $\boldsymbol{\sigma}^t_{{\mbtheta}_{i,0}} \gets \boldsymbol{\sigma}^t_{{\mbtheta}}$
        \FOR{$k=0,1, \dots, K-1$}
        \STATE Update the MAP values over representation parameters ${\mbtheta}^t_{i,\mathrm{MAP}} \gets \mathrm{GD}(\boldsymbol{\mu}^t_{{\mbtheta}_{i,0}}, \boldsymbol{\sigma}^t_{{\mbtheta}_{i,0}}, {\mbphi}^0_i)$
        \ENDFOR
        \STATE Determine the subnetwork ${\mbtheta}_{i,S}^t$ as Eq.~\eqref{17}
        \STATE Infer the full-covariance Gaussian posterior over the subnetwork $q({\mbtheta}^t_{i,S})$ as Eq.~\eqref{18}
        
        \textbf{Return} $\boldsymbol{\mu}^{t+1}_{{\mbtheta}_i}$, $\boldsymbol{\sigma}^{t+1}_{{\mbtheta}_i}$
	\end{algorithmic}  
\end{algorithm}

\section{Experiments}
\label{section 5}

\subsection{Experimental Setup}
\textbf{Datasets and Models.} We compare FedSI with several relevant baselines on three popular benchmark datasets: MNIST \cite{lecun1998gradient}, FMNIST \cite{xiao2017fashion}, and CIFAR-10 \cite{krizhevsky2009learning}. MNIST and FMNIST contain 10 labels and 70,000 images of handwritten digits and clothing instances, with 60,000 for training and 10,000 for test. CIFAR-10 comprises of 60,000 RGB images with 10 labels, with 50,000 for training and 10,000 for test. Due to the distinction in data sizes among these datasets, we set the number of clients to 10 for the MNIST and FMNIST datasets, while the number of clients for the CIFAR-10 dataset is set to 20. To model a heterogeneous setting associated with clients, we assign 5 of 10 labels to each client. Similar to \cite{collins2021exploiting} and \cite{zhang2022personalized}, we use a multilayer perceptron (MLP) with a single hidden layer for the MNIST and FMNIST datasets. For the CIFAR-10 dataset, we employ a 5-layer convolutional neural network (CNN), which consists of two convolutional layers followed by three fully connected layers.

\textbf{Baselines.} We compare FedSI with the following baselines.

\begin{enumerate}
\item Local only, each client trains its own model locally, and there is no communication with other clients throughout the entire training process. 

\item FedAvg \cite{mcmahan2017communication}, a standard FL model for learning a single global model.

\item FedAvg-FT, a locally fine-tuning version of FedAvg.

\item FedBABU \cite{oh2021fedbabu}, a DNNs-based PFL method that only updates and aggregates the common representation layers during the entire federated training process, while the decision layer is obtained by fine-tuning.



\item FedPer \cite{arivazhagan2019federated}, a DNNs-based PFL method that learns the common representation layers and the personalized decision layer through simultaneous local updating.

\item FedRep \cite{collins2021exploiting}, a DNNs-based PFL method  similar to  FedPer but with alternating local updating.

\item LG-FedAvg \cite{liang2020think}, a DNNs-based PFL method that learns the common decision layer and personalized representation layers.

\item pFedBayes \cite{zhang2022personalized}, a Bayesian PFL method that approximates the posterior distribution over model parameters through MFVI and achieves the personalization using hyperparameter $\zeta$.

\item BPFed \cite{chen2023bayesian}, a Bayesian DNNs-based PFL method that learns the common representation layers and the personalized decision layer in a continual learning fashion, and approximates the posterior distribution over entire model parameters through MFVI.

\item FedPop \cite{kotelevskii2022fedpop}, a Bayesian DNNs-based PFL method that learns the common representation layers with deterministic model parameters and the personalized decision layer with stochastic model parameters whose posterior distribution is approximated by SGLD.

\item FedSI-Fac, a variant of FedSI for verifying the efficacy of subnetwork inference, which approximates the posterior distribution over the representation layers with a factorized assumption (i.e., diagonal Hessian matrix).
\end{enumerate}

\textbf{Implementation Details.} On each communication round, we fix the random subset of all clients $C=10$ for all three benchmark datasets. The number of communication rounds is set to 800 on all three datasets, which is motivated by the observation that additional communication rounds do not significantly contribute to accuracy improvement for all FL approaches. We perform 10 epochs to train the local model and employ  Adam \cite{kingma2014adam} as the local optimizer for all FL algorithms. Additionally, we run all experiments in this paper on a system comprising 4 cores, with each core powered by an Intel(R) Xeon(R) CPU E5-2686 at a frequency of 2.30GHz. The system was further enhanced by the inclusion of a NVIDIA Tesla K80 GPU. In addition to the GPU memory, the system is boasted by a total of 60GB of memory.

\textbf{Hyperparameters.} For the proposed FedSI, we set the prior variance $\alpha$ = 1e-4 which demonstrates favorable performance in our experiments. Moreover, we tune the learning rate over \{1e-3, 5e-3, 1e-2, 5e-2, 0.1\} and fix it to 1e-2. For pFedBayes, we set the hyperparameter $\zeta=10$ and the learning rate to 1e-4. For other algorithms, we set the learning rate to 1e-3. Note that FedSI-Fac shares the same hyperparameter setting with FedSI. For all algorithms, we set the batch size to 50 on all datasets. 

\subsection{Effect of Size of Subnetworks} 
\begin{figure}[h]
    \centering
    \begin{subfigure}[b]{0.5\textwidth}
        \includegraphics[width=\textwidth]{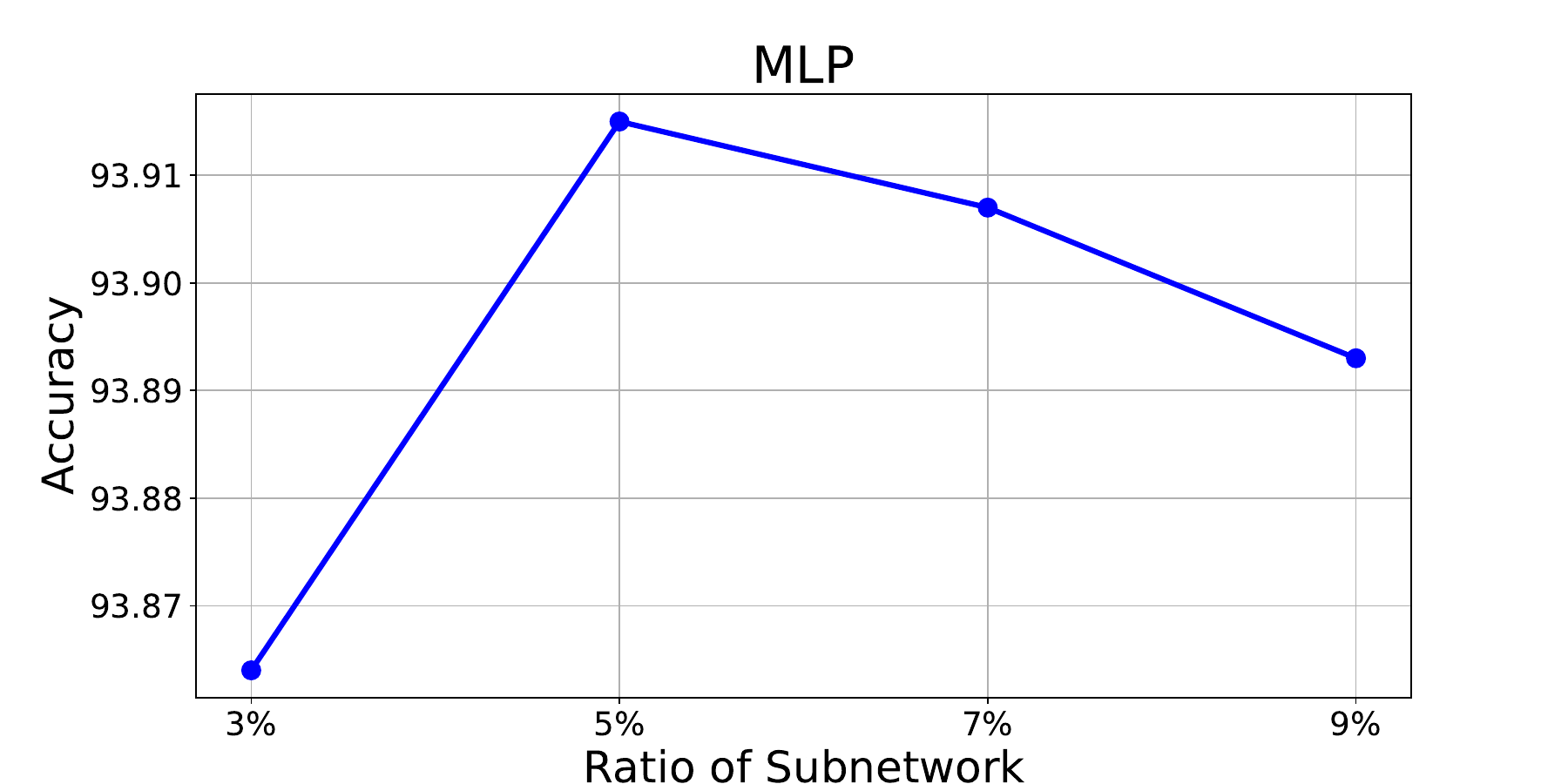}
        \label{fig:cifar10_accuracy}
    \end{subfigure}
    \begin{subfigure}[b]{0.5\textwidth}
        \includegraphics[width=\textwidth]{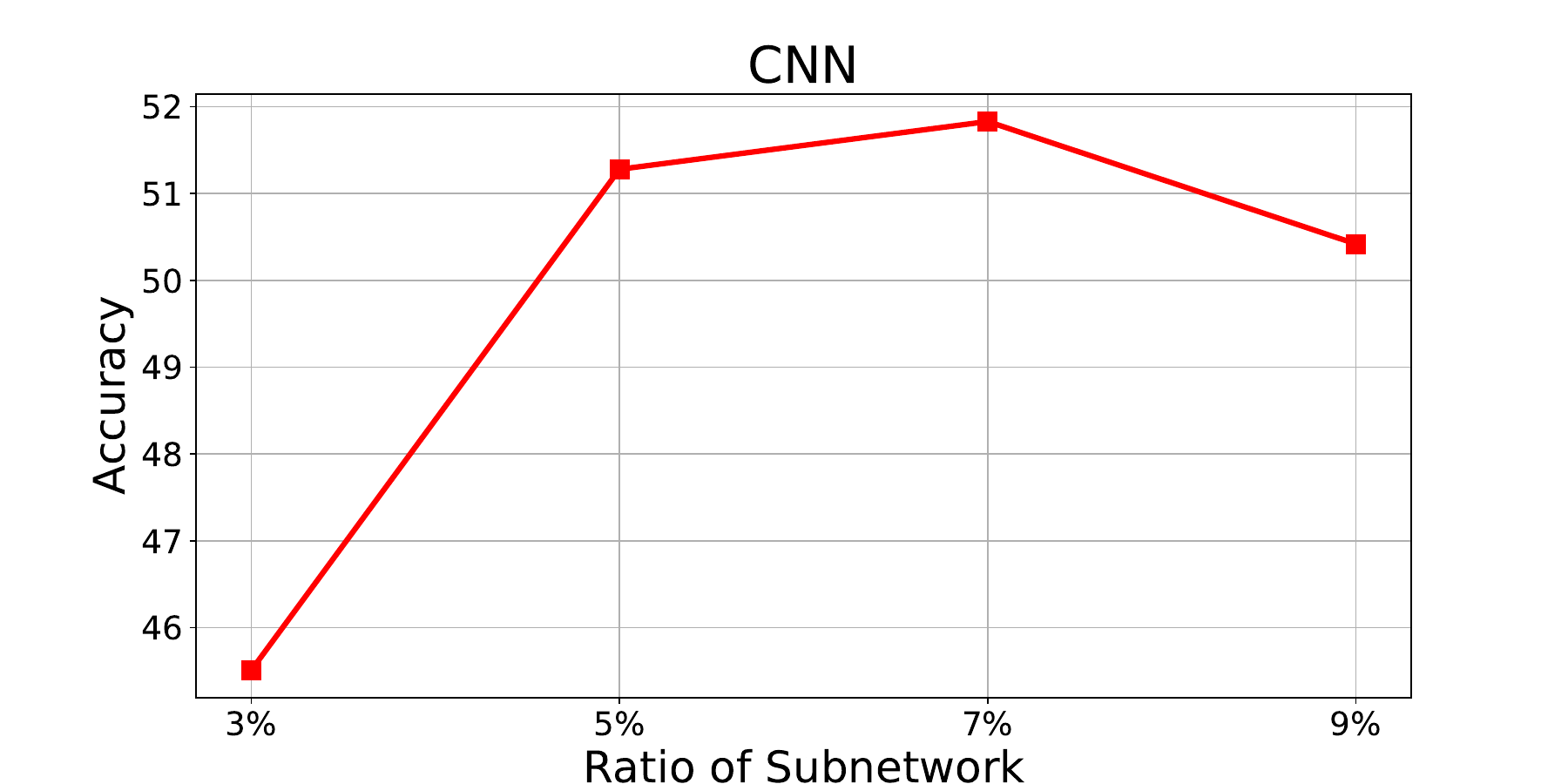}
        \label{fig:mnist_accuracy}
    \end{subfigure}
    \caption{Test accuracy comparison with varying ratios of the subnetwork for MLP and CNN.}
    \label{fig3}
\end{figure}

For FedSI, the number of parameters in the subnetwork $S$ is considered a hyperparameter. In order to understand the effect of this hyperparameter on the performance of different model structures (i.e., MLP and CNN), we conduct a series of experiments on the MNIST and CIFAR-10 datasets. It is important to note that, as the number of subnetwork's parameters increases, the model tends to retain more uncertainty, but concurrently incurs additional computational costs. Therefore, we aim to select an appropriate number of subnetwork's parameters while considering model performance. We denote the ratio of the number of subnetwork's parameters to the total number of parameters in the representation layers as $E$ and tune $E \in \{3\%, 5\%, 7\%, 9\%\}$ for both  MLP and CNN. In Fig. \ref{fig3}, we observe that the MLP model achieves its best performance on the MNIST dataset when $E=5\%$, while the CNN model attains its optimal performance on the CIFAR-10 dataset when $E=7\%$. Since these two hyperparameter settings achieve the best model performance while preserving sufficient uncertainty, we set $E=5\%$ and $E=7\%$ for the MLP and CNN models separately in the remaining experiments.

\subsection{Effect of Data Size on Classification} 
In order to evaluate the effect of data size on model performance, we conduct the experiments on all three datasets and consider the similar data partition as in \cite{zhang2022personalized}. Specifically, we arrange small and large subsets separately for each of three benchmark datasets. For the small and large subsets of MNIST and FMNIST, we utilize 50 training samples and 950 test samples for each class in the small subset; while in the large subset, we use 900 training samples and 300 testing samples for each class. For the small and large subsets of CIFAR-10, we employ 25 training samples and 475 test samples for each class in the small subset; and in the large subset, we allocate 450 training samples and 150 testing samples for each class. We report the average test accuracy across clients. The main results are presented in Table \ref{tab1}, we can see that our method performs the best on all three small subsets, which demonstrates the superiority of our designed Bayesian framework on limited data. On the other hand, it is obvious that FedSI consistently outperforms  FedSI-Fac on all datasets, revealing that the factorized assumption leads to a degradation in model performance. Interestingly, despite having a simple algorithmic framework, FedAvg-FT shows strong competitiveness in comparison with other start-of-the-art algorithms, especially on small datasets. 

\begin{table*}[t]
\renewcommand{\arraystretch}{1.3}
\caption{The comparison of average test accuracy (\%) on MNIST, FMNIST and CIFAR-10 w.r.t. small and large data size. 
  }
\label{tab1}
\centering
\begin{tabular}{ccccccc}
\toprule
Dataset & \multicolumn{2}{c}{MNIST} & \multicolumn{2}{c}{FMNIST} & \multicolumn{2}{c}{CIFAR-10} \\

\cmidrule(lr){2-3} \cmidrule(lr){4-5} \cmidrule(lr){6-7}
Dataset size                              & Small & Large & Small & Large & Small & Large \\
 \midrule
Local Only & $91.38\pm0.02$ & $95.16\pm0.05$ & $86.51\pm0.02$ & $90.98\pm0.07$ & $40.98\pm0.75$ & $71.10\pm0.57$ \\ 
FedAvg & $88.26\pm0.01$ & $89.54\pm0.03$ & $81.28\pm0.02$ & $82.54\pm0.07$ & $36.51\pm0.05$ & $58.15\pm0.07$ \\
FedAvg-FT & $93.15\pm0.02$ & $94.42\pm0.03$ & $88.05\pm0.03$ & $89.67\pm0.06$ & $48.93\pm0.37$ & $76.11\pm0.82$ \\

 \midrule
 FedBABU  & $91.38\pm0.06$ & $98.26\pm0.03$ & $87.70\pm0.11$ & $93.66\pm0.06$ & $49.05\pm0.03$& $\boldsymbol{82.78\pm0.72}$ \\ 
 FedPer & $91.82\pm0.07$ & $98.49\pm0.02$ & $86.62\pm0.03$ & $96.69\pm0.01$ & $48.16\pm1.23$ & $80.90\pm0.16$ \\
 FedRep & $90.68\pm0.09$ & $98.58\pm0.02$ & $86.13\pm0.07$ & $\boldsymbol{96.78\pm0.02}$ & $46.17\pm0.60$ & $82.51\pm0.26$ \\
 LG-FedAvg & $91.64\pm0.05$ & $97.13\pm0.03$ & $86.35\pm0.05$ & $95.66\pm0.03$ & $42.15\pm0.49$ & $71.19\pm0.24$  \\
 
\midrule 
 pFedBayes & $84.61\pm0.06$ & $92.54\pm0.04$ & $76.74\pm0.18$ & $84.31\pm0.22$ & $48.24\pm1.55$ & $70.23\pm0.48$ \\
 BPFed & $92.96\pm0.38$ & $98.43\pm0.01$ & $86.54\pm0.17$ & $91.76\pm0.24$ & $49.79\pm1.22$ & $68.05\pm0.06$ \\
 FedPop & $87.63\pm0.38$ & $89.35\pm0.23$ & $83.75\pm0.99$ & $82.17\pm0.03$ & $43.64\pm1.24$  & $51.29\pm0.25$ \\
 FedSI-Fac & $91.96\pm0.02$ & $97.96\pm0.03$ & $86.55\pm0.06$ & $94.41\pm0.03$ & $49.63\pm0.62$ & $74.61\pm0.17$ \\
\midrule 
FedSI (Ours) & $\boldsymbol{93.98\pm0.05 }$ & $\boldsymbol{99.28\pm0.02}$ & $\boldsymbol{88.41\pm0.02}$ & $96.36\pm0.03$ & $\boldsymbol{51.21\pm1.22}$ & $76.1\pm0.87$ \\
\bottomrule
\end{tabular}
\end{table*} 

\subsection{Main Results on Uncertainty Quantification} 
\begin{figure*}[h]
    \centering
    \begin{subfigure}[b]{0.3\textwidth}
        \includegraphics[width=\textwidth]{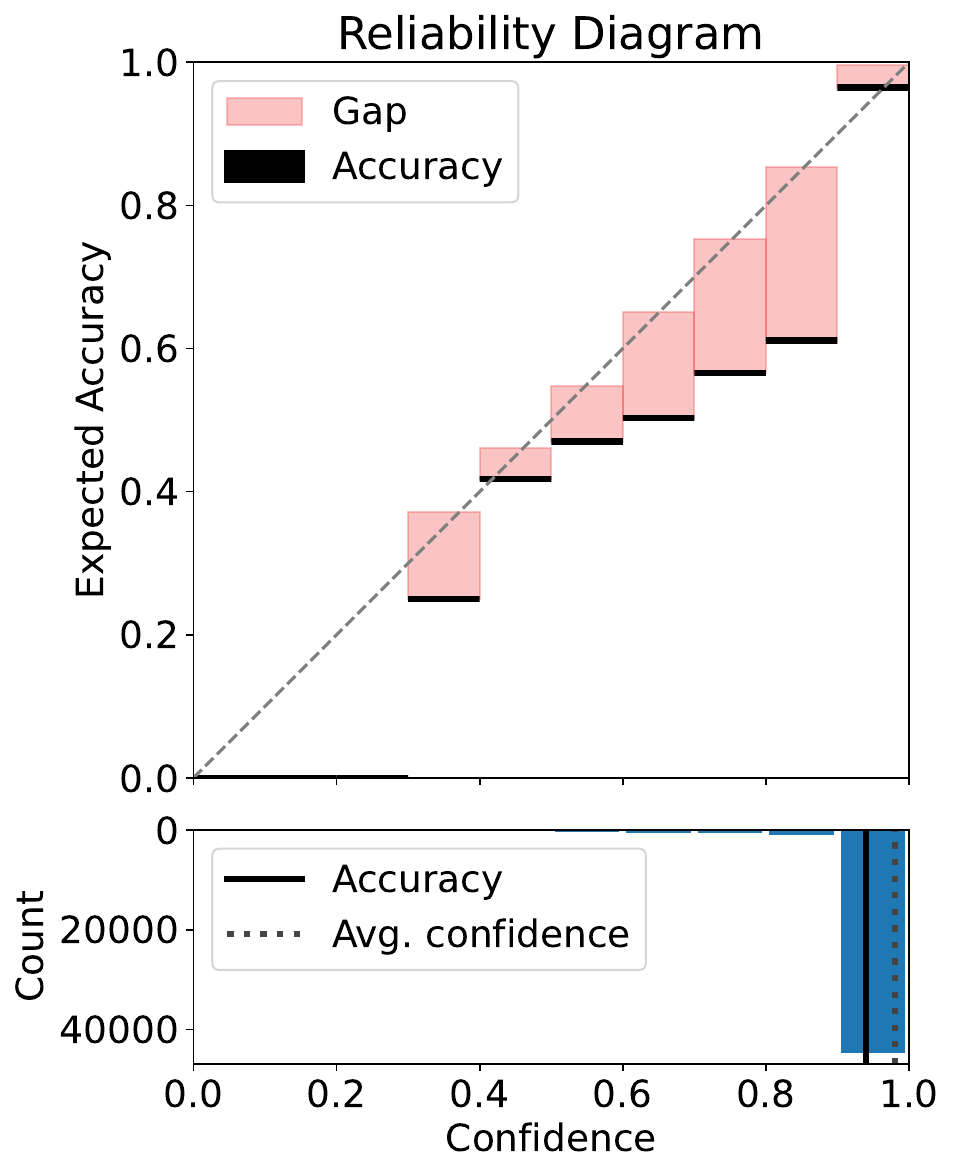}
        \label{fig:cifar10_accuracy}
    \end{subfigure}
    \begin{subfigure}[b]{0.3\textwidth}
        \includegraphics[width=\textwidth]{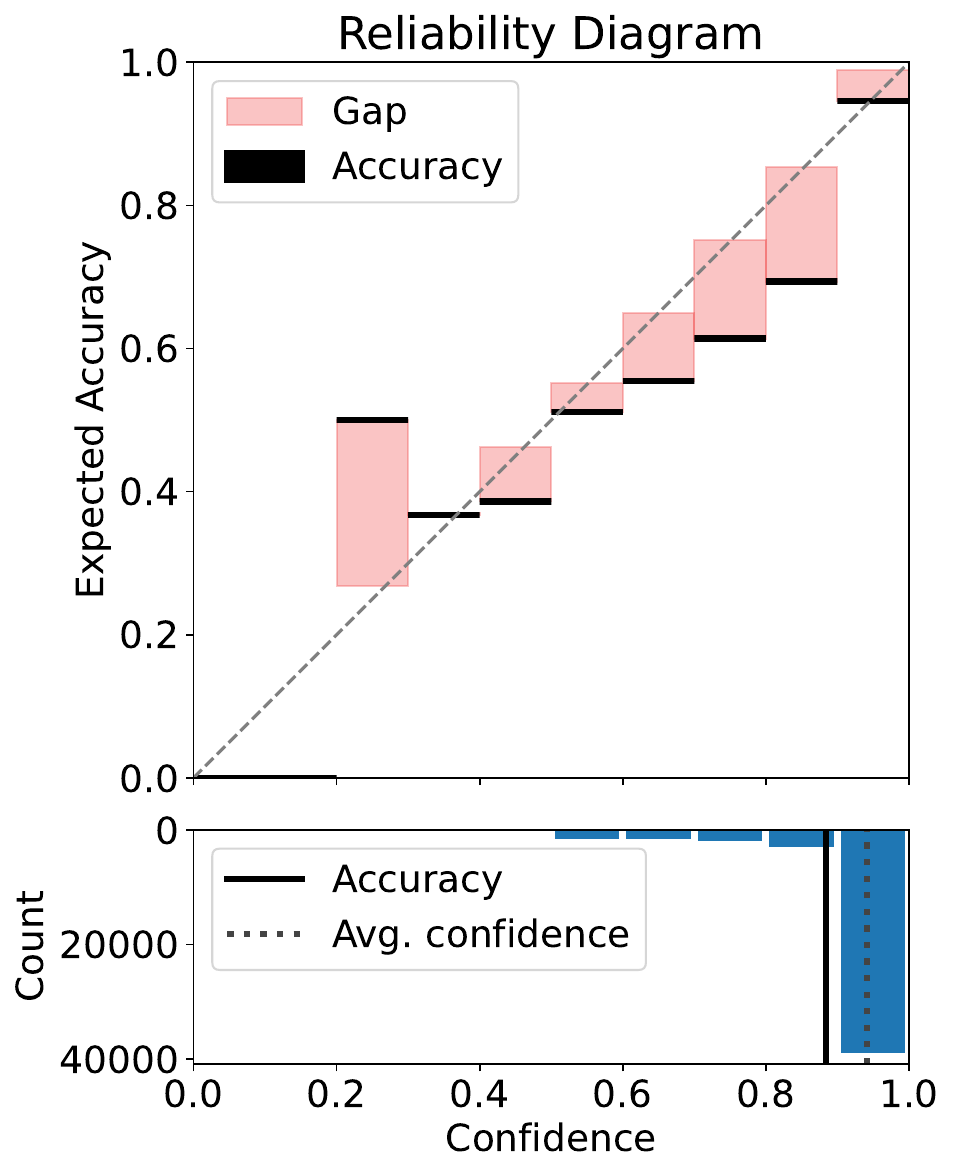}
        \label{fig:mnist_accuracy}
    \end{subfigure}
    \begin{subfigure}[b]{0.3\textwidth}
        \includegraphics[width=\textwidth]{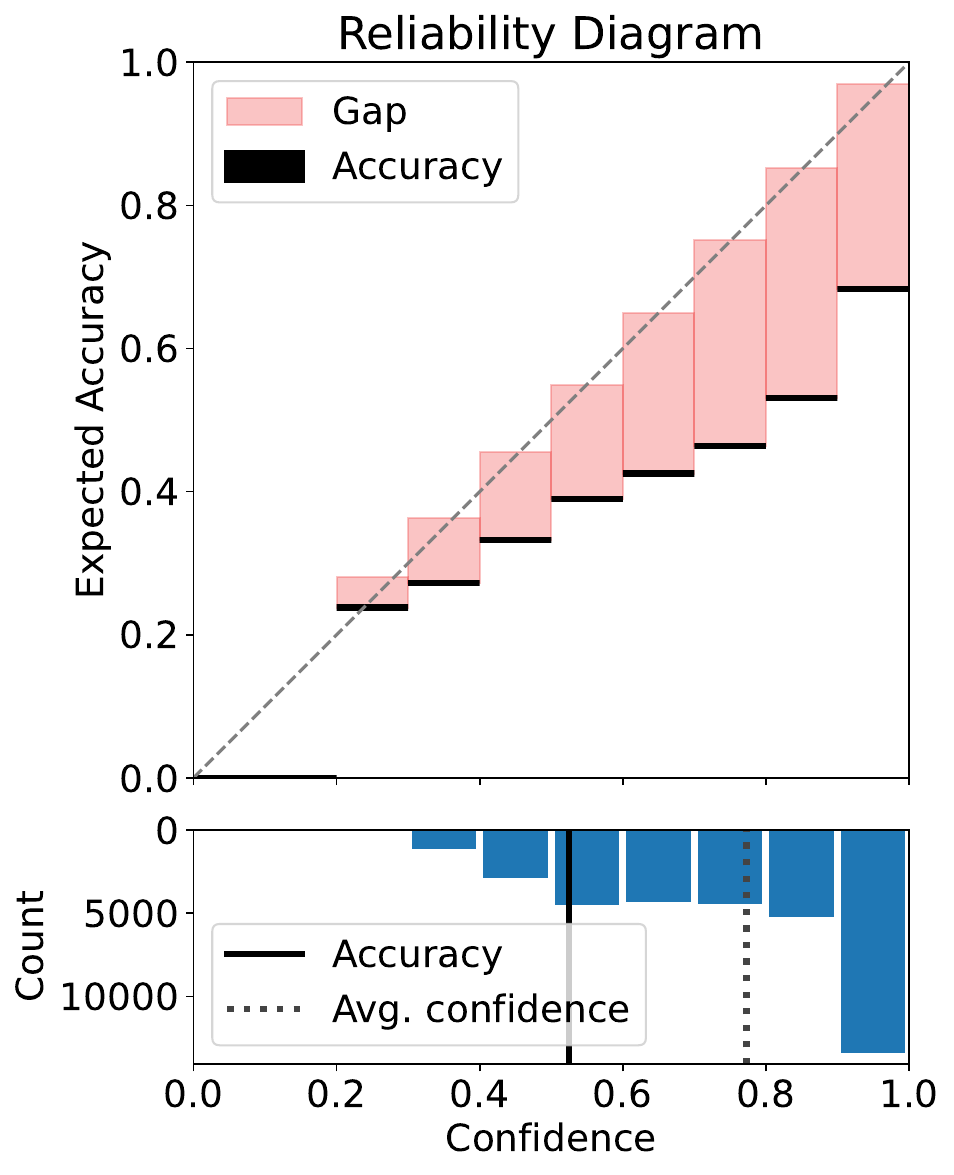}
        \label{fig:cifar10_accuracy}
    \end{subfigure}
    \caption{Reliability diagram and confidence histogram of FedSI on MNIST (left), FMNIST (middle) and CIFAR-10 (right). The closer the accuracy line and the average confidence line are, the better the model is calibrated.}
    \label{fig4}
\end{figure*}

By introducing stochastic parameters into the FL model, our method has the capability to perform uncertainty quantification, which is a crucial property in some domains such as autonomous driving, healthcare. Here, we illustrate this property through conducting confidence calibration experiments on three benchmark datasets. Fig. \ref{fig4} shows the calibration results on three benchmark datasets, we can find that our method is well-calibrated on all three datasets, especially on MNIST and FMNIST. 

To further explore the uncertainty calibration performance of the proposed method, we conduct experiments on three benchmark datasets and compare FedSI with other Bayesian FL methods. We choose three metrics \cite{guo2017calibration}: the Expectation Calibration Error (ECE) calculates the weighted average of absolute difference between the predicted average confidence and the actual accuracy, the Maximum Calibration Error (MCE) focuses on the maximum discrepancy between predicted probabilities and the actual accuracy, and the BRIer score (BRI) measures the mean squared difference between predicted probabilities and the actual labels, to evaluate the uncertainty calibration performance. The numerical results are shown in Table \ref{tab2}, from which we can see that our method outperforms other Bayesian FL baselines over the metrics of ECE and MCE on all three datasets. For the metric of BRI, FedSI still achieves the best performance on FMNIST and CIFAR-10 datasets while loses to the pFedBayes by a small margin on the MNIST dataset. In particular, the general improvement from FedSI-Fac to FedSI across all datasets on the three metrics indicates that the performance of uncertainty calibration benefits from the subnetwork inference strategy.

 


\begin{table*}[t]
\renewcommand{\arraystretch}{1.3}
\caption{The comparison of different calibration metrics on MNIST, FMNIST and CIFAR-10. 
  }
\label{tab2}
\centering
\begin{tabular}{cccccccccc}
\toprule
Dataset & \multicolumn{3}{c}{MNIST} & \multicolumn{3}{c}{FMNIST} & \multicolumn{3}{c}{CIFAR-10} \\

\cmidrule(lr){2-4} \cmidrule(lr){5-7} \cmidrule(lr){8-10}
Metric  & ECE & MCE & BRI & ECE & MCE & BRI & ECE & MCE & BRI\\
 \midrule
pFedBayes & $0.046$ & $0.257$ & $\boldsymbol{0.100}$ & $0.062$ & $0.287$ & $0.190$ & $0.456$ & $0.444$ & $0.832 $\\
BPFed & $0.043$ & $0.291$ & $0.123$ & $0.093$ & $0.299$ & $0.209$ & $0.431$ & $0.502 $ & $0.803$\\
FedPop & $0.049$ & $0.716$ & $0.137$ & $0.103$ & $0.280$  & $0.237$ & $0.470$ & $0.514$ & $0.874$\\
FedSI-Fac & $0.045$ & $0.249$ & $0.131$ & $0.068$ & $0.283$ & $0.281$ & $0.290$ & $0.358$ & $0.741$\\
\midrule 
FedSI (Ours) & $\boldsymbol{0.040}$ & $\boldsymbol{0.242}$ & $0.101$ & $\boldsymbol{0.057}$ & $\boldsymbol{0.232}$ & $\boldsymbol{0.182}$ & $\boldsymbol{0.249}$ & $\boldsymbol{0.321}$ & $\boldsymbol{0.704}$\\
\bottomrule
\end{tabular}
\end{table*} 

\subsection{Generalization to Novel Clients} 
\begin{figure*}[h]
    \centering
    \begin{subfigure}[b]{0.3\textwidth}
        \includegraphics[width=\textwidth]{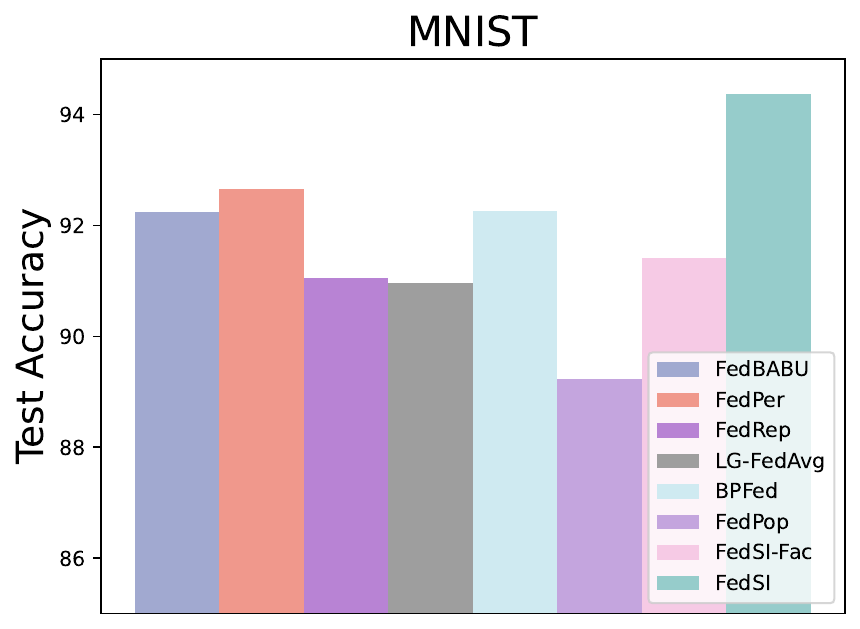}
        \label{fig:cifar10_accuracy}
    \end{subfigure}
    \begin{subfigure}[b]{0.3\textwidth}
        \includegraphics[width=\textwidth]{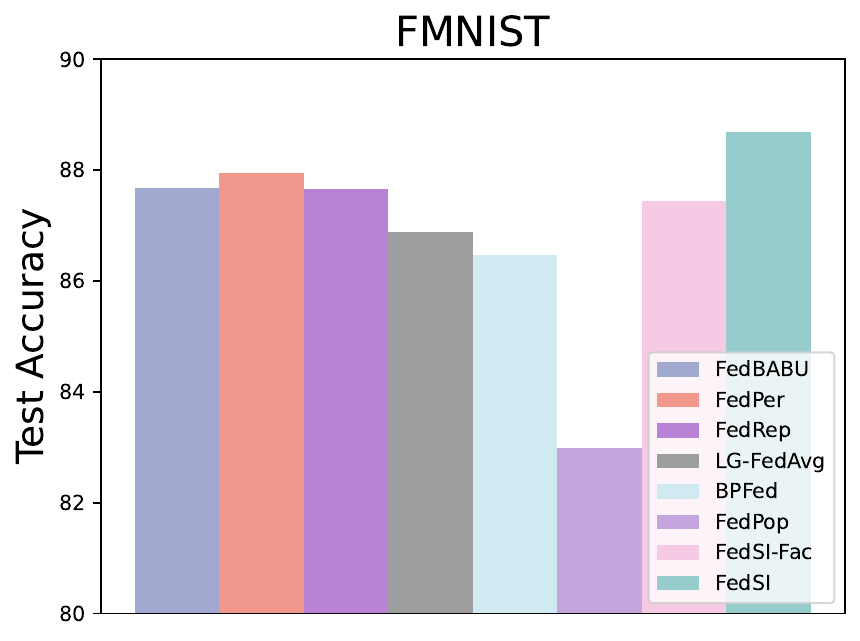}
        \label{fig:mnist_accuracy}
    \end{subfigure}
    \begin{subfigure}[b]{0.3\textwidth}
        \includegraphics[width=\textwidth]{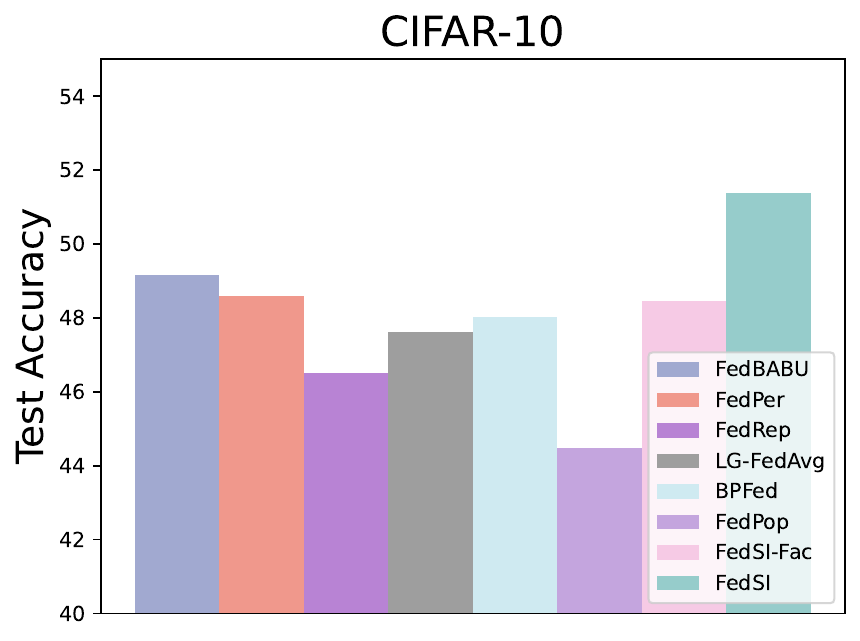}
        \label{fig:cifar10_accuracy}
    \end{subfigure}
    \caption{Performance comparison of different DNNs-based PFL algorithms on MNIST, FMNIST and CIFAR-10.}
    \label{fig5}
\end{figure*}

FL is a dynamic system, often involving novel clients which enter the system after model training completion. This presents a significant challenge for traditional FL frameworks as they require retraining all model parameters for  novel clients. In contrast, akin to other DNNs-based PFL algorithms, FedSI is capable of executing Step 4 in Section \ref{section 3} to generalize to  novel clients easily  with low computational cost. We consider similar experimental protocol in \cite{collins2021exploiting} and compare FedSI with other DNNs-based PFL baselines on three benchmark datasets. In the case of each dataset, we partition the clients into two segments: the novel client for model test, and the remaining clients for model training. During model training, the goal is to derive the common representation layers, which are subsequently used in conjunction with a novel client's samples as inputs to acquire low-dimensional personalized decision layer. The results in Fig. \ref{fig5} show that our method outperforms other baselines on all three benchmark datasets, especially on MNIST and CIFAR-10, which verifies its stability and adaptability in dynamic and heterogeneous FL settings. We also notice that FedPop performs the worst on all three datasets, which may be attributed to negative impact of its common prior on the model performance in such dynamic and heterogeneous scenarios.

\subsection{Effect of Local Epochs}
\begin{figure}[t]
\centering
\includegraphics[width=1\linewidth]{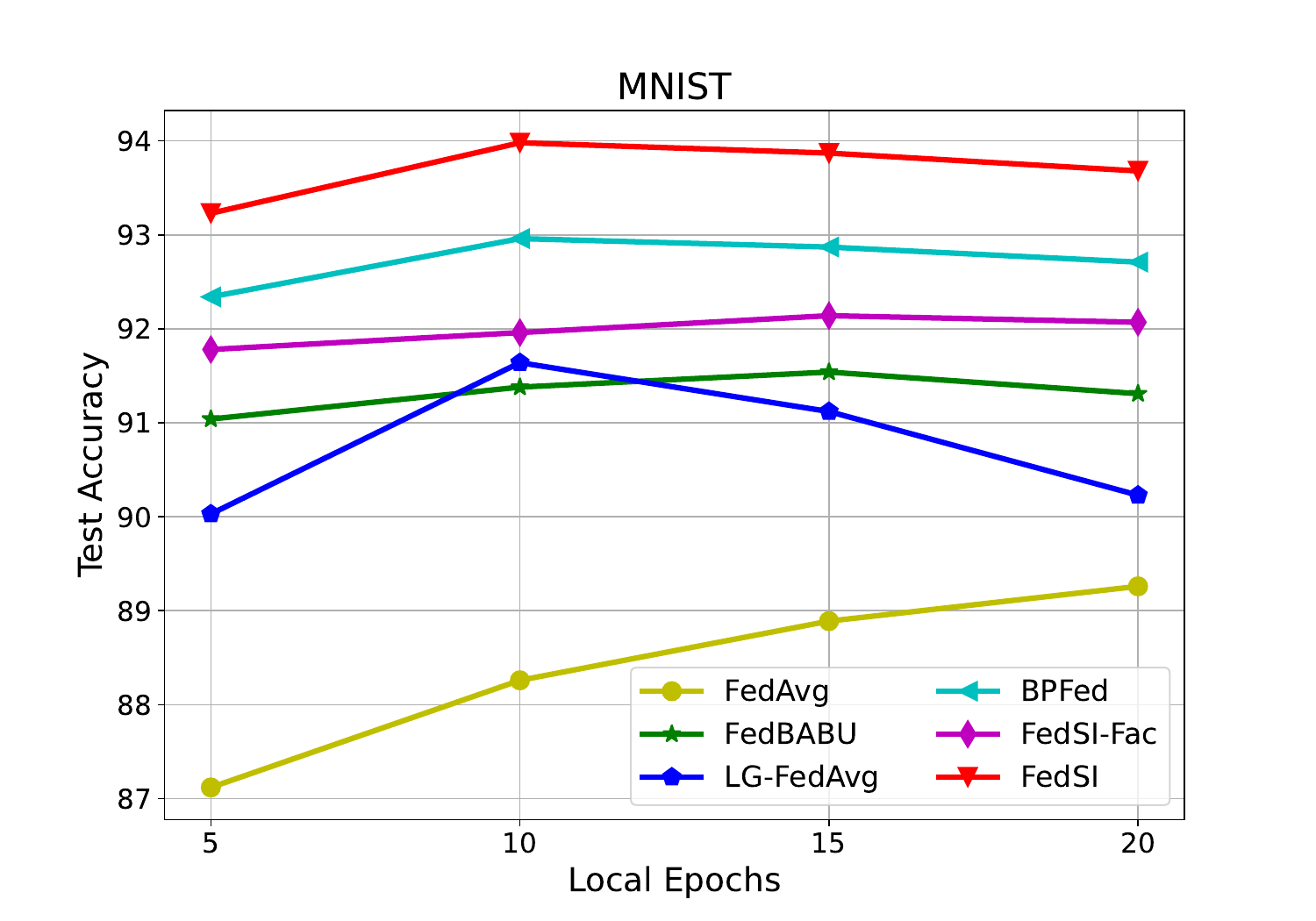}
\caption{Performance comparison on the MNIST dataset with varying local
epochs.}
\label{fig:local_epochs}
\end{figure}

Communication overhead is one of the main challenges of FL systems in practical applications. To reduce the number of communication rounds, we can increase the local epochs for each client to achieve more computation. Therefore, we investigate the effect of local epochs on model performance by varying the epochs over $\{5, 10, 15, 20\}$ and fixing the communication rounds to 800 on the MNIST dataset. The results in Fig. \ref{fig:local_epochs} conclude that our method achieves the best performance across all four local epochs. We also find that, for most algorithms, larger local epochs would lead to performance degradation, whereas the  model performance of FedAvg could still benefit from an increase of local epochs. Moreover, FedSI exhibits relatively smooth model performance across different local epochs, demonstrating its stability and reliability in such FL scenarios. It is noteworthy that considering the trade-off between computation and communication is crucial when selecting the appropriate local epoch for FL algorithms.

\section{Related Work}
\label{section 6}

\subsection{Bayesian Federated Learning} 
Bayesian federated learning approaches integrate Bayesian learning techniques into the federated learning framework to empower models with the ability to quantify uncertainty and enhance model stability. In general, we can categorize these methods into server-side approaches and client-side approaches \cite{cao2023bayesian}. For the server-side approaches, a Bayesian model ensemble algorithm FedBE \cite{chen2020fedbe} alleviates the performance degradation in heterogeneous data setting by treating global posterior distribution as a Gaussian or Dirichlet, but FedBE still utilizes a point estimation method to update the local model parameters instead of a BNN. Due to the strong correlation between prior and posterior, Bayesian continual learning introduced by FOLA \cite{liu2021bayesian} expedites model convergence. However, FOLA has a poor performance when local data are non-IID. By imposing a strong assumptions on uniform prior and independent local data, Al-Shedivat et al. \cite{al2020federated} proposed  FedPA to decompose the global posterior distribution into the product of local posterior distributions for addressing the statistical heterogeneity issue in local data. Based on FedPA, FedEP \cite{guo2023federated} further enhances model accuracy and convergence speed by leveraging the expectation propagation method to learn the global posterior. Unfortunately, unrealistic strong assumptions limit the applicability of these two algorithms. For the client-side approaches, both pFedBayes \cite{zhang2022personalized} and QLSD \cite{vono2022qlsd} employ BNN for local model training on each client. However, they utilize different approximate methods for posterior inference, with pFedBayes using VI and QLSD using Markov Chain Monte Carlo (MCMC). To achieve more stable performance in a heterogeneous data environment, Dai et al. \cite{dai2020federated} incorporated the federated Bayesian optimization with Thompson sampling into the FL framework for local model updates. Bayesian nonparametric methods are also used in some FL works, PFNM \cite{yurochkin2019bayesian} and FedMA \cite{wang2020federated} aim to find the matched subsets of neurons among local models based on the Beta-Bernoulli process, but both of them fail to perform well on non-IID data. pFedGP \cite{achituve2021personalized} deals with this issue by designing a personalized classifier for each client based on the Gaussian process.

\subsection{DNNs-based Personalized Federated Learning} 
By decoupling the entire DNN model into representation and decision layers, DNNs-based PFL has become a significant research branch within the PFL framework. Both FedPer \cite{arivazhagan2019federated} and FedRep \cite{collins2021exploiting} aim to learn the common representation layers and the personalized decision layer. The difference is that FedPer uses simultaneous local updates for both representation and decision layers, while FedPer uses iterative local updates. Furthermore, FedPer provides a rigorous theoretical justification. Unlike FedPer and FedRep, LG-FedAvg \cite{liang2020think} learns the common decision layer and personalized representation layer simultaneously via FedAvg. Recently, Oh et al. proposed FedBABU to investigate the relationship between the decoupled components and model performance by learning the common representation layers while keeping the decision layer fixed throughout the entire federated training process. However, the above methods lack ability for uncertainty quantification and perform poorly on small-scale datasets. To address these issues, FedPop \cite{kotelevskii2022fedpop} introduces the uncertainty over the parameters of personalized decision layer and learns the common representation layers across the clients. Compared to  FedSI, FedPop is unsuitable for heterogeneous structures across clients due to the common prior over the personalized decision layer and the substantial computational and memory costs caused by its approximation method SGLD are significant limitations. Recently, Chen et al. \cite{chen2023bayesian} proposed a general Bayesian PFL algorithm BPFed by introducing uncertainty over the entire BNN for learning common representation layers and personalized decision layer. Different from our method, BPFed utilizes MFVI to approximate the posterior, which hurts the model performance, especially on complex datasets.

\section{Conclusion}
\label{section 7}
Existing Bayesian DNNs-based PFL algorithms often rely on complete factorization assumptions or extensive sampling for posterior approximation over model parameters, which either severely harm model performance or result in significant computational and memory costs. In this paper, we propose a novel Bayesian DNNs-based PFL framework and algorithm for efficient uncertainty quantification in the federated setting. By introducing the subnetwork inference and fine-tuning over the representation layers and decision layer separately, FedSI can not only capture the uncertainty in prediction but also lead to improved model performance over a variety of FL baselines. In the future, efficient uncertainty quantification in the context of federated learning will remain our research focus, including new frameworks based on alternative approximation methods or different model structures.


\appendix
\section*{Derivation for GGN Approximation}
For the Laplace approximation described in Section \ref{section 2-B}, we obtain the approximated posterior distribution:
\begin{equation}
\begin{aligned}
&\log p(\mbw |\mathcal{D}) \\
&\simeq \log p(\mbw_{\mathrm{MAP}} | \mathcal{D})-\frac{1}{2}(\mbw-\mbw_{\mathrm{MAP}})^{\mathrm{T}} \mathbf{H}(\mbw-\mbw_{\mathrm{MAP}}),
\end{aligned}
\end{equation}
where the Hessian matrix $\mathbf{H}\in \mathbb{R}^{P \times P}$ is defined by
\begin{equation}
\label{eq:22}
\begin{aligned}
\mathbf{H}&=-\left.\nabla^2_{\mbw\mbw} \log p(\mbw|\mathcal{D})\right|_{\mbw=\mbw_{\mathrm{MAP}}}\\
&=- \nabla^2_{\mbw\mbw} \log p(\mathbf{y} | f_{\mbw}(\mathbf{X}))+\alpha^{-1} \mathbf{I} |_{\mbw=\mbw_{\mathrm{MAP}}}.
\end{aligned}
\end{equation}

Since the prior term is trivial, we focus on the log likelihood and interpret it from the perspective of a single data sample:
\begin{equation}
\label{eq:23}
\nabla^2_{\mbw\mbw}\log p(\mathbf{y} | f_{\mbw}(\mathbf{X}))=\widetilde {\mathbf{H}} \mathbf{r}-\mathbf{J}^{\mathrm{T}} \mathbf{\Lambda} \mathbf{J},
\end{equation}
where Hessian matrix $\widetilde{\mathbf{H}}=\nabla^2_{\mbw\mbw}f_{\mbw}(\mbx)$; $\mathbf{J}=\nabla_{\mbw}f_{\mbw}(\mbx)\in \mathbb{R}^{O \times P}$ denotes the Jacobian matrix of model output over the parameters. $\mathbf{r}=\nabla_{f} \log p(\mathbf{y}|f_{\mbw}(\mbx))$ and $\mathbf{\Lambda}=-\nabla^2_{ff} \log p(\mathbf{y}|f_{\mbw}(\mbx)) \in \mathbb{R}^{O \times O}$ represent the residual and noise precision, respectively.

In practice, we cannot compute the term $\widetilde {\mathbf{H}}$ in Eq.~\eqref{eq:23}, but we can use GGN to drop this term for approximating Eq.~\eqref{eq:22} as:
\begin{equation}
\label{eq:GGN}
\mathbf{H} \simeq \widehat{\mathbf{H}}= \mathbf{J}^{\mathrm{T}} \mathbf{\Lambda} \mathbf{J}+\alpha^{-1} \mathbf{I}.
\end{equation}

\section*{Derivation for Eq.~\eqref{16}}
For the squared 2-Wasserstein distance between two Gaussian distributions $\mathcal{N}(\boldsymbol{\mu}_1,\boldsymbol{\Sigma}_1)$ and $\mathcal{N}(\boldsymbol{\mu}_2,\boldsymbol{\Sigma}_2)$, we  give the closed-form expression as follows:
\begin{equation}
\begin{aligned}
&W_2\left(\mathcal{N}\left(\boldsymbol{\mu}_1, \boldsymbol{\Sigma}_1\right), \mathcal{N}\left(\boldsymbol{\mu}_2, \boldsymbol{\Sigma}_2\right)\right)^2\\
&=\left\|\boldsymbol{\mu}_1-\boldsymbol{\mu}_2\right\|_2^2+\operatorname{Tr}\left(\boldsymbol{\Sigma}_1+\boldsymbol{\Sigma}_2-2\left(\boldsymbol{\Sigma}_2^{1 / 2} \boldsymbol{\Sigma}_1 \boldsymbol{\Sigma}_2^{1 / 2}\right)^{1 / 2}\right).    
\end{aligned}
\end{equation}

In this paper, the true posterior $p({\mbtheta}_i|\mathcal{D}_i)$ and approximate posterior from subnetwrok $q_S({\mbtheta}_{i})$ share the same mean ${\mbtheta}_{i,\mathrm{MAP}}$ (i.e., $\boldsymbol{\mu}_1=\boldsymbol{\mu}_2={\mbtheta}_{i,\mathrm{MAP}}$). The covariance matrix of true posterior $\boldsymbol{\Sigma}_1$ is the inverse GGN matrix $\widehat{\mathbf{H}}_i^{-1}$, while the covariance matrix of approximate posterior from subnetwork can be regarded as the inverse GGN matrix $\widehat{\mathbf{H}}_{i,S}^{-1}$ with zeros at the positions of ${\mbtheta}_{i,D}$ for matching the shape of $\widehat{\mathbf{H}}_i^{-1}$ (i.e., $\boldsymbol{\Sigma}_2=\widehat{\mathbf{H}}_{i,S^*}^{-1}$). Thus, the squared 2-Wasserstein distance between $p({\mbtheta}_i|\mathcal{D}_i)$ and $q_S({\mbtheta}_{i})$ is obtained by
\begin{equation}
\begin{aligned}
& W_2\left(p({\mbtheta}_i|\mathcal{D}_i), q_S({\mbtheta}_{i})\right)^2 \\
& =W_2\left(\mathcal{N}\left({\mbtheta}_{i,\mathrm{MAP}}, \widehat{\mathbf{H}}_i^{-1}\right), \mathcal{N}\left({\mbtheta}_{i,\mathrm{MAP}}, \widehat{\mathbf{H}}_{i,S^*}^{-1}\right)\right)^2 \\
& =||{\mbtheta}_{i,\mathrm{MAP}}-{\mbtheta}_{i,\mathrm{MAP}} ||_2^2\\
&+\operatorname{Tr}\left(\widehat{\mathbf{H}}_i^{-1}+\widehat{\mathbf{H}}_{i,S^*}^{-1}-2\sqrt{\widehat{\mathbf{H}}_{i,S}^{-1/2} \widehat{\mathbf{H}}_i^{-1} \widehat{\mathbf{H}}_{i,S^*}^{-1/2}}\right) \\
& =\operatorname{Tr}\left(\widehat{\mathbf{H}}_i^{-1}+\widehat{\mathbf{H}}_{i,S^*}^{-1}-2\sqrt{\widehat{\mathbf{H}}_{i,S}^{-1/2} \widehat{\mathbf{H}}_i^{-1} \widehat{\mathbf{H}}_{i,S^*}^{-1/2}}\right) .
\end{aligned}
\end{equation}

\section*{Derivation for Eq.~\eqref{17}}
By giving an independent assumption over the model parameters, we have $\widehat{\mathbf{H}}_i^{-1}=\mathrm{diag}(\sigma_1,\cdots,\sigma_{r})$. Eq.~\eqref{16} can be simplified to
\begin{equation}
\begin{aligned}
& W_2\left(p({\mbtheta}_i|\mathcal{D}_i), q_S({\mbtheta}_{i})\right)^2 \\
&=\operatorname{Tr}\left(\widehat{\mathbf{H}}_i^{-1}\right)+\operatorname{Tr}\left(\widehat{\mathbf{H}}_{i,S^*}^{-1}\right)-2 \operatorname{Tr}\left(\widehat{\mathbf{H}}_i^{-1/2} \widehat{\mathbf{H}}_{i,S^*}^{-1 / 2}\right) \\
&=\sum_{r=1}^R \sigma_r^2+\sigma_r^2 \chi_{{\mbtheta}_{i,S}} ({\mbtheta}_{i,r})-2 \sigma_r^2 \chi_{{\mbtheta}_{i,S}} ({\mbtheta}_{i,r}) \\
& =\sum_{r=1}^R \sigma_d^2\left(1-\chi_{{\mbtheta}_{i,S}} ({\mbtheta}_{i,r})\right),
\end{aligned}
\end{equation}
where $\chi$ is the indicator function.



\bibliography{TNNLS}

\begin{thebibliography}{10}

\bibitem{cao2023bayesian}
L.~Cao, H.~Chen, X.~Fan, J.~Gama, Y.-S. Ong, and V.~Kumar, ``Bayesian federated
  learning: A survey,'' {\em IJCAI}, 2023.

\bibitem{li2022evolutionary}
T.~Li, L.~Chen, C.~S. Jensen, T.~B. Pedersen, Y.~Gao, and J.~Hu, ``Evolutionary
  clustering of moving objects,'' in {\em ICDE}, pp.~2399--2411, IEEE, 2022.

\bibitem{guo2023gpt4graph}
J.~Guo, L.~Du, and H.~Liu, ``Gpt4graph: Can large language models understand
  graph structured data? an empirical evaluation and benchmarking,'' {\em arXiv
  preprint arXiv:2305.15066}, 2023.

\bibitem{zhao2022personalized}
Y.~Zhao, G.~Yu, J.~Wang, C.~Domeniconi, M.~Guo, X.~Zhang, and L.~Cui,
  ``Personalized federated few-shot learning,'' {\em IEEE Transactions on
  Neural Networks and Learning Systems}, 2022.

\bibitem{tan2022towards}
A.~Z. Tan, H.~Yu, L.~Cui, and Q.~Yang, ``Towards personalized federated
  learning,'' {\em IEEE Transactions on Neural Networks and Learning Systems},
  2022.

\bibitem{Cao_DeAI}
L.~Cao, ``Decentralized ai: Edge intelligence and smart blockchain, metaverse,
  web3, and desci,'' {\em IEEE Intelligent Systems}, vol.~37, no.~03,
  pp.~6--19, 2022.

\bibitem{mcmahan2017communication}
B.~McMahan, E.~Moore, D.~Ramage, S.~Hampson, and B.~A. y~Arcas,
  ``Communication-efficient learning of deep networks from decentralized
  data,'' in {\em AISTATS}, pp.~1273--1282, PMLR, 2017.

\bibitem{t2020personalized}
C.~T~Dinh, N.~Tran, and J.~Nguyen, ``Personalized federated learning with
  moreau envelopes,'' {\em NeurIPS}, vol.~33, pp.~21394--21405, 2020.

\bibitem{achituve2021personalized}
I.~Achituve, A.~Shamsian, A.~Navon, G.~Chechik, and E.~Fetaya, ``Personalized
  federated learning with gaussian processes,'' {\em NeurIPS}, vol.~34,
  pp.~8392--8406, 2021.

\bibitem{yao2022trajgat}
D.~Yao, H.~Hu, L.~Du, G.~Cong, S.~Han, and J.~Bi, ``Trajgat: A graph-based
  long-term dependency modeling approach for trajectory similarity
  computation,'' in {\em KDD}, pp.~2275--2285, 2022.

\bibitem{du2022understanding}
L.~Du, X.~Chen, F.~Gao, Q.~Fu, K.~Xie, S.~Han, and D.~Zhang, ``Understanding
  and improvement of adversarial training for network embedding from an
  optimization perspective,'' in {\em WSDM}, pp.~230--240, 2022.

\bibitem{C22beyiid}
L.~Cao, ``Beyond i.i.d.: Non-iid thinking, informatics, and learning,'' {\em
  {IEEE} Intell. Syst.}, vol.~37, no.~4, pp.~5--17, 2022.

\bibitem{chen2021bridging}
H.-Y. Chen and W.-L. Chao, ``On bridging generic and personalized federated
  learning for image classification,'' {\em ICLR}, 2022.

\bibitem{xu2023personalized}
J.~Xu, X.~Tong, and S.-L. Huang, ``Personalized federated learning with feature
  alignment and classifier collaboration,'' {\em ICLR}, 2023.

\bibitem{li2020compression}
T.~Li, R.~Huang, L.~Chen, C.~S. Jensen, and T.~B. Pedersen, ``Compression of
  uncertain trajectories in road networks,'' {\em VLDB}, vol.~13, no.~7,
  pp.~1050--1063, 2020.

\bibitem{li2021trace}
T.~Li, L.~Chen, C.~S. Jensen, and T.~B. Pedersen, ``Trace: Real-time
  compression of streaming trajectories in road networks,'' {\em VLDB},
  vol.~14, no.~7, pp.~1175--1187, 2021.

\bibitem{arivazhagan2019federated}
M.~G. Arivazhagan, V.~Aggarwal, A.~K. Singh, and S.~Choudhary, ``Federated
  learning with personalization layers,'' {\em arXiv preprint
  arXiv:1912.00818}, 2019.

\bibitem{collins2021exploiting}
L.~Collins, H.~Hassani, A.~Mokhtari, and S.~Shakkottai, ``Exploiting shared
  representations for personalized federated learning,'' in {\em ICML},
  pp.~2089--2099, PMLR, 2021.

\bibitem{oh2021fedbabu}
J.~Oh, S.~Kim, and S.-Y. Yun, ``Fedbabu: Towards enhanced representation for
  federated image classification,'' {\em ICLR}, 2022.

\bibitem{cheng2017survey}
Y.~Cheng, D.~Wang, P.~Zhou, and T.~Zhang, ``A survey of model compression and
  acceleration for deep neural networks,'' {\em IEEE Signal Processing
  Magazine}, 2017.

\bibitem{smith2017federated}
V.~Smith, C.-K. Chiang, M.~Sanjabi, and A.~S. Talwalkar, ``Federated multi-task
  learning,'' {\em NeurIPS}, vol.~30, 2017.

\bibitem{fallah2020personalized}
A.~Fallah, A.~Mokhtari, and A.~Ozdaglar, ``Personalized federated learning with
  theoretical guarantees: A model-agnostic meta-learning approach,'' {\em
  NeurIPS}, vol.~33, pp.~3557--3568, 2020.

\bibitem{amodei2016concrete}
D.~Amodei, C.~Olah, J.~Steinhardt, P.~Christiano, J.~Schulman, and D.~Man{\'e},
  ``Concrete problems in ai safety,'' {\em arXiv preprint arXiv:1606.06565},
  2016.

\bibitem{kokolakis2022safety}
N.-M.~T. Kokolakis and K.~G. Vamvoudakis, ``Safety-aware pursuit-evasion games
  in unknown environments using gaussian processes and finite-time convergent
  reinforcement learning,'' {\em IEEE Transactions on Neural Networks and
  Learning Systems}, 2022.

\bibitem{liu2023dynamic}
Z.~Liu and X.~He, ``Dynamic submodular-based learning strategy in imbalanced
  drifting streams for real-time safety assessment in nonstationary
  environments,'' {\em IEEE Transactions on Neural Networks and Learning
  Systems}, 2023.

\bibitem{jiang2024reinforcement}
K.~Jiang, Z.~Jiang, X.~Jiang, Y.~Xie, and W.~Gui, ``Reinforcement learning for
  blast furnace ironmaking operation with safety and partial observation
  considerations,'' {\em IEEE Transactions on Neural Networks and Learning
  Systems}, 2024.

\bibitem{kairouz2021advances}
P.~Kairouz, H.~B. McMahan, B.~Avent, A.~Bellet, M.~Bennis, A.~N. Bhagoji,
  K.~Bonawitz, Z.~Charles, G.~Cormode, R.~Cummings, {\em et~al.}, ``Advances
  and open problems in federated learning,'' {\em Foundations and
  Trends{\textregistered} in Machine Learning}, vol.~14, no.~1--2, pp.~1--210,
  2021.

\bibitem{du2022gbk}
L.~Du, X.~Shi, Q.~Fu, X.~Ma, H.~Liu, S.~Han, and D.~Zhang, ``Gbk-gnn: Gated
  bi-kernel graph neural networks for modeling both homophily and
  heterophily,'' in {\em Proceedings of the ACM Web Conference 2022},
  pp.~1550--1558, 2022.

\bibitem{liu2024probabilistic}
H.~Liu, T.~Zhang, F.~Li, M.~Yu, and G.~Yu, ``A probabilistic generative model
  for tracking multi-knowledge concept mastery probability,'' {\em Frontiers of
  Computer Science}, vol.~18, no.~3, p.~183602, 2024.

\bibitem{yu2023subspace}
M.~Yu, X.~Chen, X.~Gu, H.~Liu, and L.~Du, ``A subspace constraint based
  approach for fast hierarchical graph embedding,'' {\em World Wide Web},
  vol.~26, no.~5, pp.~3691--3705, 2023.

\bibitem{welling2011bayesian}
M.~Welling and Y.~W. Teh, ``Bayesian learning via stochastic gradient langevin
  dynamics,'' in {\em ICML}, pp.~681--688, 2011.

\bibitem{fan2021continuous}
X.~Fan, B.~Li, F.~Zhou, and S.~SIsson, ``Continuous-time edge modelling using
  non-parametric point processes,'' {\em NeurIPS}, vol.~34, pp.~2319--2330,
  2021.

\bibitem{fan2021poisson}
X.~Fan, B.~Li, Y.~Li, and S.~A. Sisson, ``Poisson-randomised dirbn: large
  mutation is needed in dirichlet belief networks,'' in {\em ICML},
  pp.~3068--3077, PMLR, 2021.

\bibitem{rodriguez2022function}
S.~Rodriguez-Santana, B.~Zaldivar, and D.~Hernandez-Lobato, ``Function-space
  inference with sparse implicit processes,'' in {\em ICML}, pp.~18723--18740,
  PMLR, 2022.

\bibitem{fan2023free}
X.~Fan, E.~V. Bonilla, T.~O’Kane, and S.~A. Sisson, ``Free-form variational
  inference for gaussian process state-space models,'' in {\em ICML},
  pp.~9603--9622, PMLR, 2023.

\bibitem{fan2020bayesian}
X.~Fan, B.~Li, L.~Luo, and S.~A. Sisson, ``Bayesian nonparametric space
  partitions: A survey,'' {\em arXiv preprint arXiv:2002.11394}, 2020.

\bibitem{wu2024weakly}
Z.~Wu, L.~Cao, Q.~Zhang, J.~Zhou, and H.~Chen, ``Weakly augmented variational
  autoencoder in time series anomaly detection,'' {\em arXiv preprint
  arXiv:2401.03341}, 2024.

\bibitem{izmailov2020subspace}
P.~Izmailov, W.~J. Maddox, P.~Kirichenko, T.~Garipov, D.~Vetrov, and A.~G.
  Wilson, ``Subspace inference for bayesian deep learning,'' in {\em UAI},
  pp.~1169--1179, PMLR, 2020.

\bibitem{sharma2023bayesian}
M.~Sharma, S.~Farquhar, E.~Nalisnick, and T.~Rainforth, ``Do bayesian neural
  networks need to be fully stochastic?,'' in {\em AISTATS}, pp.~7694--7722,
  PMLR, 2023.

\bibitem{daxberger2021bayesian}
E.~Daxberger, E.~Nalisnick, J.~U. Allingham, J.~Antor{\'a}n, and J.~M.
  Hern{\'a}ndez-Lobato, ``Bayesian deep learning via subnetwork inference,'' in
  {\em ICML}, pp.~2510--2521, PMLR, 2021.

\bibitem{pillutla2022federated}
K.~Pillutla, K.~Malik, A.-R. Mohamed, M.~Rabbat, M.~Sanjabi, and L.~Xiao,
  ``Federated learning with partial model personalization,'' in {\em ICML},
  pp.~17716--17758, PMLR, 2022.

\bibitem{kotelevskii2022fedpop}
N.~Kotelevskii, M.~Vono, A.~Durmus, and E.~Moulines, ``Fedpop: A bayesian
  approach for personalised federated learning,'' {\em NeurIPS}, vol.~35,
  pp.~8687--8701, 2022.

\bibitem{foong2019between}
A.~Y. Foong, Y.~Li, J.~M. Hern{\'a}ndez-Lobato, and R.~E. Turner,
  ``'in-between'uncertainty in bayesian neural networks,'' {\em ICML 2019
  Workshop on Uncertainty and Robustness in Deep Learning}, 2019.

\bibitem{immer2021improving}
A.~Immer, M.~Korzepa, and M.~Bauer, ``Improving predictions of bayesian neural
  nets via local linearization,'' in {\em AISTATS}, pp.~703--711, PMLR, 2021.

\bibitem{bishop2006pattern}
C.~M. Bishop and N.~M. Nasrabadi, {\em Pattern recognition and machine
  learning}, vol.~4.
\newblock Springer, 2006.

\bibitem{martens2020new}
J.~Martens, ``New insights and perspectives on the natural gradient method,''
  {\em The Journal of Machine Learning Research}, vol.~21, no.~1,
  pp.~5776--5851, 2020.

\bibitem{barber1998ensemble}
D.~Barber and C.~M. Bishop, ``Ensemble learning in bayesian neural networks,''
  {\em Nato ASI Series F Computer and Systems Sciences}, vol.~168,
  pp.~215--238, 1998.

\bibitem{fan2018rectangular}
X.~Fan, B.~Li, and S.~Sisson, ``Rectangular bounding process,'' {\em NeurIPS},
  vol.~31, 2018.

\bibitem{ovadia2019can}
Y.~Ovadia, E.~Fertig, J.~Ren, Z.~Nado, D.~Sculley, S.~Nowozin, J.~Dillon,
  B.~Lakshminarayanan, and J.~Snoek, ``Can you trust your model's uncertainty?
  evaluating predictive uncertainty under dataset shift,'' {\em NeurIPS},
  vol.~32, 2019.

\bibitem{fort2019deep}
S.~Fort, H.~Hu, and B.~Lakshminarayanan, ``Deep ensembles: A loss landscape
  perspective,'' {\em arXiv preprint arXiv:1912.02757}, 2019.

\bibitem{foong2020expressiveness}
A.~Foong, D.~Burt, Y.~Li, and R.~Turner, ``On the expressiveness of approximate
  inference in bayesian neural networks,'' {\em NeurIPS}, vol.~33,
  pp.~15897--15908, 2020.

\bibitem{lecun2015deep}
Y.~LeCun, Y.~Bengio, and G.~Hinton, ``Deep learning,'' {\em nature}, vol.~521,
  no.~7553, pp.~436--444, 2015.

\bibitem{chizat2020faster}
L.~Chizat, P.~Roussillon, F.~L{\'e}ger, F.-X. Vialard, and G.~Peyr{\'e},
  ``Faster wasserstein distance estimation with the sinkhorn divergence,'' {\em
  NeurIPS}, vol.~33, pp.~2257--2269, 2020.

\bibitem{lecun1998gradient}
Y.~LeCun, L.~Bottou, Y.~Bengio, and P.~Haffner, ``Gradient-based learning
  applied to document recognition,'' {\em Proceedings of the IEEE}, vol.~86,
  no.~11, pp.~2278--2324, 1998.

\bibitem{xiao2017fashion}
H.~Xiao, K.~Rasul, and R.~Vollgraf, ``Fashion-mnist: a novel image dataset for
  benchmarking machine learning algorithms,'' {\em arXiv preprint
  arXiv:1708.07747}, 2017.

\bibitem{krizhevsky2009learning}
A.~Krizhevsky, ``Learning multiple layers of features from tiny images,'' {\em
  Master's thesis, University of Tront}, 2009.

\bibitem{zhang2022personalized}
X.~Zhang, Y.~Li, W.~Li, K.~Guo, and Y.~Shao, ``Personalized federated learning
  via variational bayesian inference,'' in {\em ICML}, pp.~26293--26310, PMLR,
  2022.

\bibitem{liang2020think}
P.~P. Liang, T.~Liu, L.~Ziyin, N.~B. Allen, R.~P. Auerbach, D.~Brent,
  R.~Salakhutdinov, and L.-P. Morency, ``Think locally, act globally: Federated
  learning with local and global representations,'' {\em NeurIPS Workshop on
  Federated Learning}, 2020.

\bibitem{chen2023bayesian}
H.~Chen, H.~Liu, L.~Cao, and T.~Zhang, ``Bayesian personalized federated
  learning with shared and personalized uncertainty representations,'' {\em
  arXiv preprint arXiv:2309.15499}, 2023.

\bibitem{kingma2014adam}
D.~P. Kingma and J.~Ba, ``Adam: A method for stochastic optimization,'' {\em
  arXiv preprint arXiv:1412.6980}, 2014.

\bibitem{guo2017calibration}
C.~Guo, G.~Pleiss, Y.~Sun, and K.~Q. Weinberger, ``On calibration of modern
  neural networks,'' in {\em ICML}, pp.~1321--1330, PMLR, 2017.

\bibitem{chen2020fedbe}
H.-Y. Chen and W.-L. Chao, ``Fedbe: Making bayesian model ensemble applicable
  to federated learning,'' {\em ICLR}, 2021.

\bibitem{liu2021bayesian}
L.~Liu, F.~Zheng, H.~Chen, G.-J. Qi, H.~Huang, and L.~Shao, ``A bayesian
  federated learning framework with online laplace approximation,'' {\em arXiv
  preprint arXiv:2102.01936}, 2021.

\bibitem{al2020federated}
M.~Al-Shedivat, J.~Gillenwater, E.~Xing, and A.~Rostamizadeh, ``Federated
  learning via posterior averaging: A new perspective and practical
  algorithms,'' {\em ICLR}, 2021.

\bibitem{guo2023federated}
H.~Guo, P.~Greengard, H.~Wang, A.~Gelman, Y.~Kim, and E.~P. Xing, ``Federated
  learning as variational inference: A scalable expectation propagation
  approach,'' {\em ICLR}, 2023.

\bibitem{vono2022qlsd}
M.~Vono, V.~Plassier, A.~Durmus, A.~Dieuleveut, and E.~Moulines, ``Qlsd:
  Quantised langevin stochastic dynamics for bayesian federated learning,'' in
  {\em AISTATS}, pp.~6459--6500, PMLR, 2022.

\bibitem{dai2020federated}
Z.~Dai, B.~K.~H. Low, and P.~Jaillet, ``Federated bayesian optimization via
  thompson sampling,'' {\em NeurIPS}, vol.~33, pp.~9687--9699, 2020.

\bibitem{yurochkin2019bayesian}
M.~Yurochkin, M.~Agarwal, S.~Ghosh, K.~Greenewald, N.~Hoang, and Y.~Khazaeni,
  ``Bayesian nonparametric federated learning of neural networks,'' in {\em
  ICML}, pp.~7252--7261, PMLR, 2019.

\bibitem{wang2020federated}
H.~Wang, M.~Yurochkin, Y.~Sun, D.~Papailiopoulos, and Y.~Khazaeni, ``Federated
  learning with matched averaging,'' {\em ICLR}, 2020.

\end{thebibliography}
\bibliographystyle{ieeetr}

\begin{IEEEbiography}[{\includegraphics[width=1in,height=1.25in,clip,keepaspectratio]{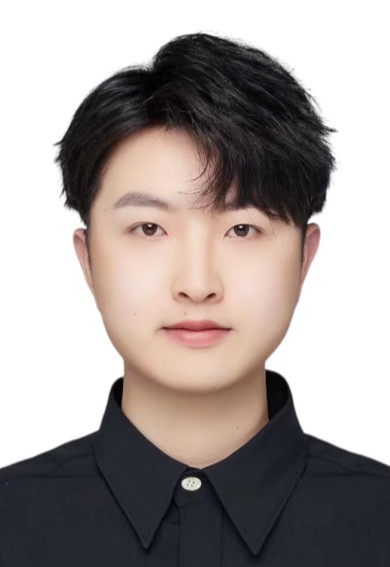}}]{Hui Chen}
is  a Ph.D. candidate in the School of Computing, Macquarie University. His current research interests include machine learning, Bayesian deep learning, federated learning and generative AI. 
\end{IEEEbiography}

\begin{IEEEbiography}[{\includegraphics[width=1in,height=1.25in,clip,keepaspectratio]{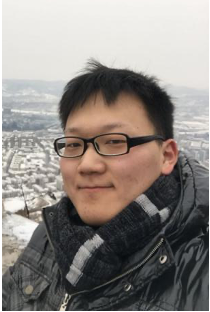}}]{Hengyu Liu} received the Ph.D. degree in computer science from Northeastern University, China, in 2023. He is  a Post-Doctoral Researcher in the Department of Computer Science at Aalborg University. His research interests are broad yet focused, encompassing Federated Learning, Education Data Management, Energy Data Management, Spatial-temporal Data Management, and Digital Twin Technology.
\end{IEEEbiography}

\begin{IEEEbiography}[{\includegraphics[width=1in,height=1.25in,clip,keepaspectratio]{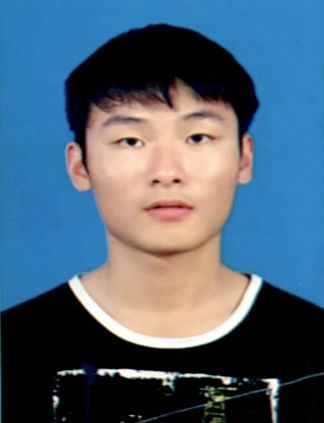}}]{Zhangkai Wu} is  a Ph.D. candidate in the Faculty of Engineering and Information Technology, University of Technology Sydney. He received his Master's degree from the Software Institute at Nanjing
University in 2020. His current research interests are deep generative models and their applications.
\end{IEEEbiography}

\begin{IEEEbiography}[{\includegraphics[width=1in,height=1.25in,clip,keepaspectratio]{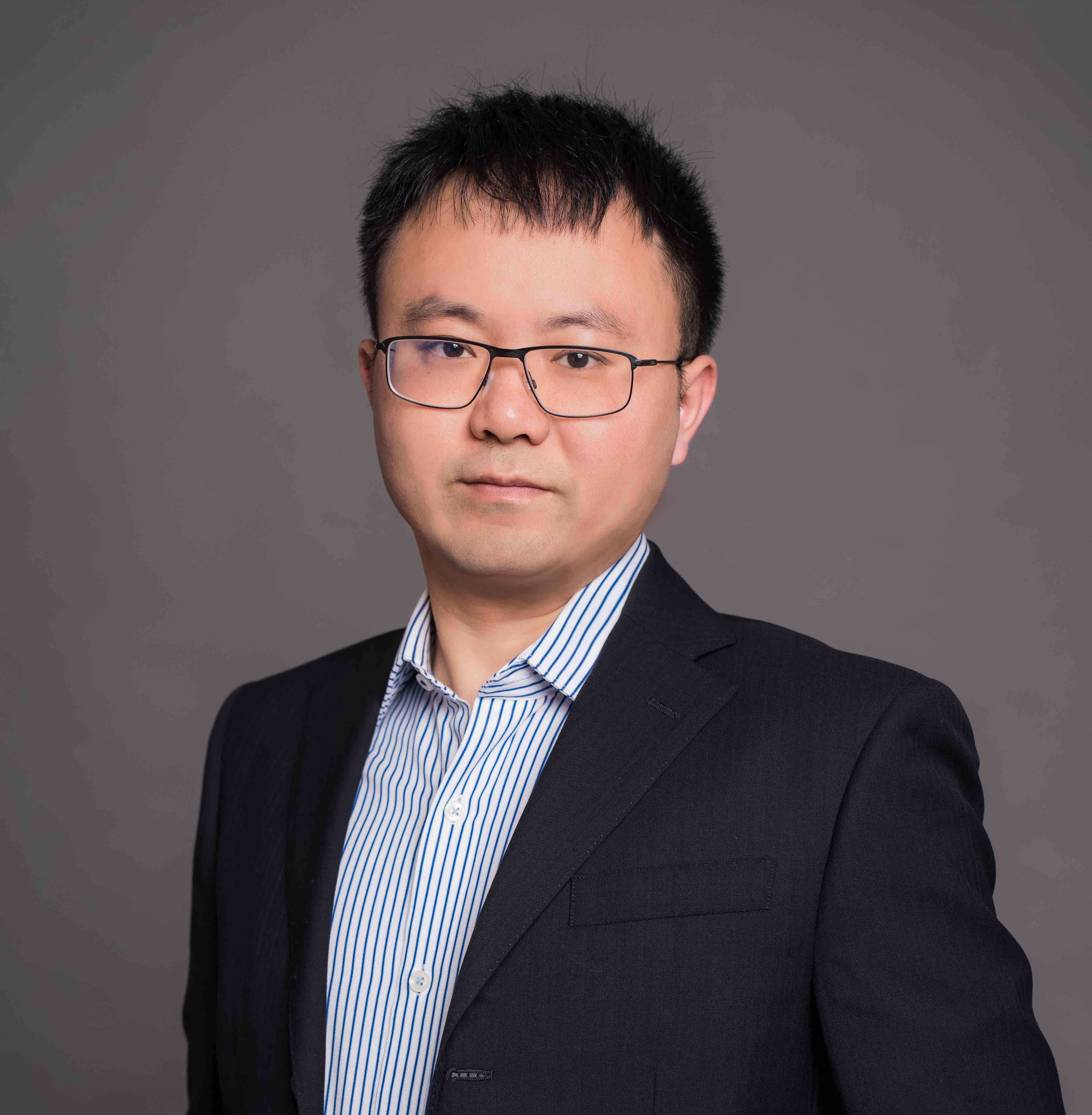}}]{Xuhui Fan} is  a lecturer in Artificial Intelligence in the School of Computing at Macquarie University, Australia. He received the bachelor's degree in mathematical statistics from the University of Science and Technology of China in 2010, and the Ph.D. degree in computer science from the University of Technology Sydney, Australia in 2015. He was a project engineer at Data61 (previously NICTA), CSIRO,  a postdoc fellow in the School of Mathematics and Statistics at the University of New South Wales, and a lecturer at the University of Newcastle, Australia. 
His current research interests include Generative AI,  Gaussian processes and their wide applications.
\end{IEEEbiography}

\begin{IEEEbiography}[{\includegraphics[width=1in,height=1.25in,clip,keepaspectratio]{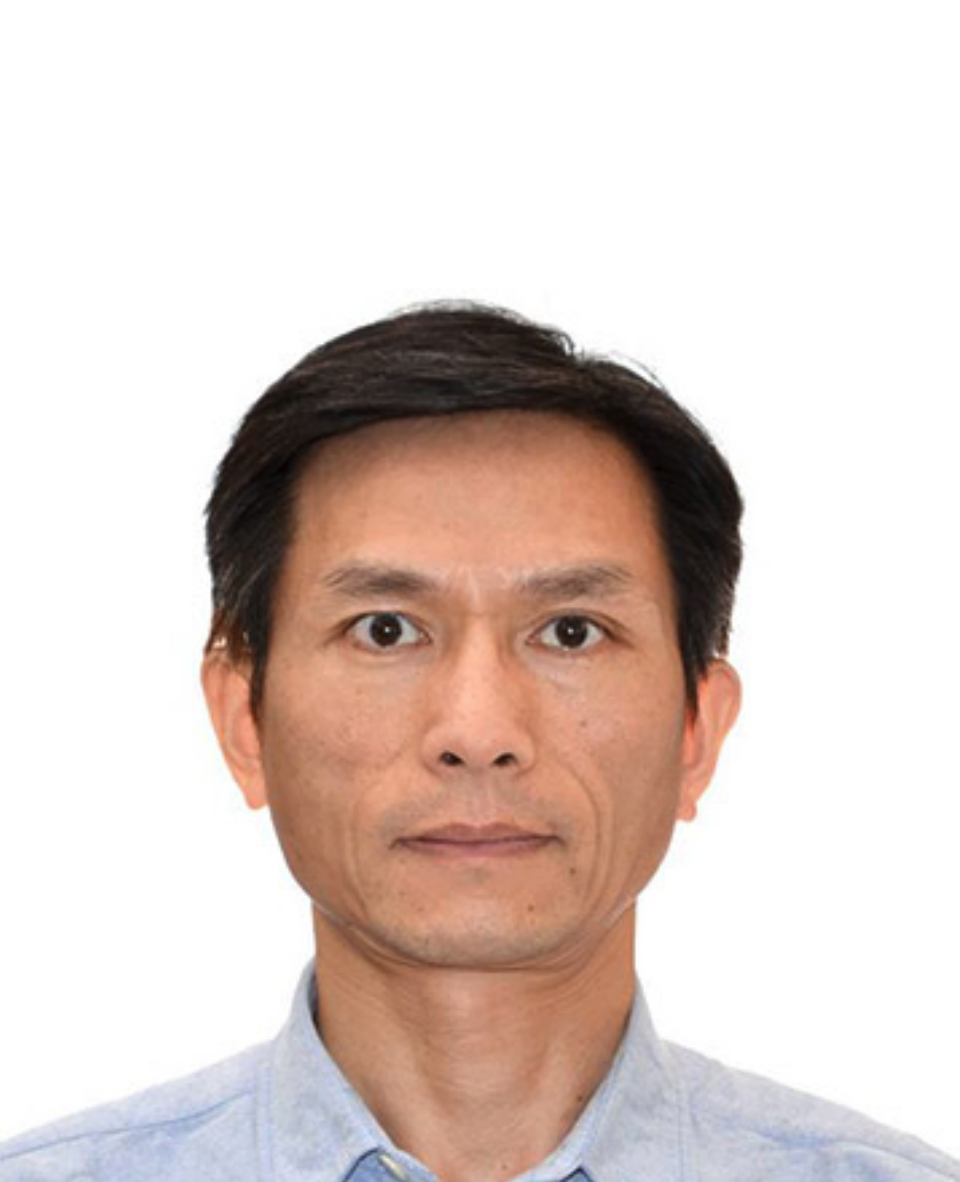}}]{Longbing Cao}(SM'06) received a PhD degree in Pattern Recognition and Intelligent Systems at Chinese Academy of Sciences in 2002 and another PhD in Computing Sciences at University of Technology Sydney in 2005. He is the Distinguished Chair Professor in AI at Macquarie University and an Australian Research Council Future Fellow (professorial level). His research interests include AI and intelligent systems, data science and analytics, machine/deep learning, behavior informatics, and enterprise innovation.
\end{IEEEbiography}

 




\vfill

\end{document}